%% file: sample-manuscript.tex
\documentclass[manuscript,screen,authorversion,nonacm]{acmart}

\AtBeginDocument{%
  \providecommand\BibTeX{{%
    \normalfont B\kern-0.5em{\scshape i\kern-0.25em b}\kern-0.8em\TeX}}}

\acmConference[]{}{}{}
\acmPrice{}
\acmISBN{}

\usepackage{xcolor}
\usepackage[flushleft]{threeparttable}
\usepackage{hyperref}
\usepackage{soul,color}
\usepackage{pdfpages}
\usepackage{array, makecell} %
\usepackage{tabularx,booktabs,siunitx}
\usepackage{subcaption}
\usepackage{pdflscape}
\usepackage{longtable}
\usepackage[inkscapeformat=png]{svg}
\usepackage{svg}

\usepackage{supertabular}
\begin{document}

%\title{Causality Inspired Explainable Artificial Intelligence for Biometrics in the Age of Deep Learning}
\title{Causality-Inspired Taxonomy for Explainable Artificial Intelligence }%in Biometrics}

% Revisiting xAI from a Causal Perspective

\author{Pedro C. Neto}
\email{pedro.d.carneiro@inesctec.pt}
\orcid{0003-1333-4889}
\affiliation{%
  \institution{INESC TEC}
  \streetaddress{R. Dr. Roberto Frias}
  \city{Porto}
  \state{Porto}
  \country{Portugal}
  \postcode{4200-465}
}
\affiliation{%
  \institution{Faculty of Engineering - University of Porto}
  \streetaddress{R. Dr. Roberto Frias}
  \city{Porto}
  \state{Porto}
  \country{Portugal}
  \postcode{4200-465}
}

\author{Tiago Gonçalves}
\affiliation{%
  \institution{INESC TEC}
  \streetaddress{R. Dr. Roberto Frias}
  \city{Porto}
  \state{Porto}
  \country{Portugal}
  \postcode{4200-465}
}
\affiliation{%
  \institution{Faculty of Engineering - University of Porto}
  \streetaddress{R. Dr. Roberto Frias}
  \city{Porto}
  \state{Porto}
  \country{Portugal}
  \postcode{4200-465}
}

\author{João Ribeiro Pinto}
\affiliation{%
  \institution{Vision-Box}
  \streetaddress{Rua Casal do Canas, no. 2}
  \city{Oeiras}
  \state{Lisboa}
  \country{Portugal}
  \postcode{2790-204}
}
% \affiliation{%
%   \institution{\textcolor{olive}{INESC TEC}}
%   \streetaddress{R. Dr. Roberto Frias}
%   \city{Porto}
%   \state{Porto}
%   \country{Portugal}
%   \postcode{4200-465}
% }
% \affiliation{%
%   \institution{\textcolor{olive}{Faculty of Engineering - University of Porto}}
%   \streetaddress{R. Dr. Roberto Frias}
%   \city{Porto}
%   \state{Porto}
%   \country{Portugal}
%   \postcode{4200-465}
% }

\author{Wilson Silva}
%\affiliation{%
%  \institution{INESC TEC}
%  \streetaddress{R. Dr. Roberto Frias}
%  \city{Porto}
%  \state{Porto}
%  \country{Portugal}
%  \postcode{4200-465}
%}
\affiliation{%
  \institution{Department of Information and Computing Sciences, and Department of Biology - Utrecht University}
  \streetaddress{Princetonplein 5}
  \city{Utrecht}
  \state{Utrecht}
  \country{The Netherlands}
  \postcode{3584 CC}
}
\affiliation{%
  \institution{Department of Radiology - The Netherlands Cancer Institute}
  \streetaddress{Plesmanlaan 121}
  \city{Amsterdam}
  \state{Noord-Holland}
  \country{The Netherlands}
  \postcode{1066 CX}
}

\author{Ana F. Sequeira}
\affiliation{%
  \institution{INESC TEC}
  \streetaddress{R. Dr. Roberto Frias}
  \city{Porto}
  \state{Porto}
  \country{Portugal}
  \postcode{4200-465}
}

\author{Arun Ross}
\affiliation{%
  \institution{Michigan State University}
  \state{Michigan}
  \country{United States}
}

\author{Jaime S. Cardoso}
\affiliation{%
  \institution{Faculty of Engineering - University of Porto}
  \streetaddress{R. Dr. Roberto Frias}
  \city{Porto}
  \state{Porto}
  \country{Portugal}
  \postcode{4200-465}
}
\affiliation{%
  \institution{INESC TEC}
  \streetaddress{R. Dr. Roberto Frias}
  \city{Porto}
  \state{Porto}
  \country{Portugal}
  \postcode{4200-465}
}

\renewcommand{\shortauthors}{Pedro C. Neto, et al.}

\begin{abstract}
  %\textcolor{red}{Systems capable of analyzing and quantifying human physical or behavioral traits, known as biometrics systems, are growing in use and application variability. Since their evolution from handcrafted features and traditional machine learning to deep learning and automatic feature extraction, the performance of biometric systems increased to outstanding values. Nonetheless, the cost of this fast progression is still not understood. Due to its opacity, deep neural networks are difficult to understand and analyze, hence, hidden capacities or decisions motivated by the wrong motives are a potential risk. Researchers have started to pivot their focus towards the understanding of deep neural networks and the explanation of their predictions. In this paper, we provide a review of the current state of explainable biometrics based on the study of 47 papers and discuss comprehensively the direction in which this field should be developed.}
 As two sides of the same coin, causality and explainable artificial intelligence (xAI) were initially proposed and developed with different goals. However, the latter can only be complete when seen through the lens of the causality framework. As such, we propose a novel causality-inspired framework for xAI that creates an environment for the development of xAI approaches. To show its applicability, biometrics was used as case study. For this, we have analysed 81 research papers on a myriad of biometric modalities and different tasks. We have categorised each of these methods according to our novel xAI Ladder and discussed the future directions of the field.
%In this paper, we propose a review of the current state of explainable biometrics through the study of 46 papers, as well as a comprehensive discussion of in which direction should this field be taken to.
\end{abstract}




\maketitle

\section{Introduction}

%%\textcolor{blue}{Topic: Total reliability on machine versus Human control of the situation(make machines more humane - explainable, interpretable, transparent, human-in-the-loop)}

Explainable Artificial Intelligence (xAI) and Causality have been two sides of the same coin for years. Over time, xAI has grown in interest and popularity due to the need to explain the ever complex artificial intelligence systems that have been developed and deployed by large companies, or research groups. Causality on the other end aimed to bring a common mathematical language to explain chains of relationships that cannot be simply described by statistics. In this document, we leverage this duality to propose a novel causality-inspired taxonomy for explainable artificial intelligence. To show its applicability, we have chosen biometrics as our case study. As such, the following document delves into biometrics and explainable artificial intelligence through the lens of causality.

Biometric systems keep growing and improving at an extremely fast pace. The SarS-COV-2 pandemic has further propelled the use of contactless biometric systems, which were already expected to be widely adopted in 2021~\cite{liubiometricscovid,pratt2021growthbio}. The real-world applications of biometric systems are quite diverse and range from face detection for virtual reality filters~\cite{Dekyvere2016} to offline smartphone authentication~\cite{Vigliarolo2020} and airport security~\cite{semedo2021contact,Youd2021}. This wide adoption implies that these systems will be used by/on individuals of varied sexes, ethnicities, and demographics overall. Hence, it is important to ensure that the systems in production are capable of handling all its potential users in an equal and fair manner. 

\begin{figure}[h!]
    \centering
    \begin{subfigure}[b]{0.35\linewidth}
       \includegraphics[width = \linewidth]{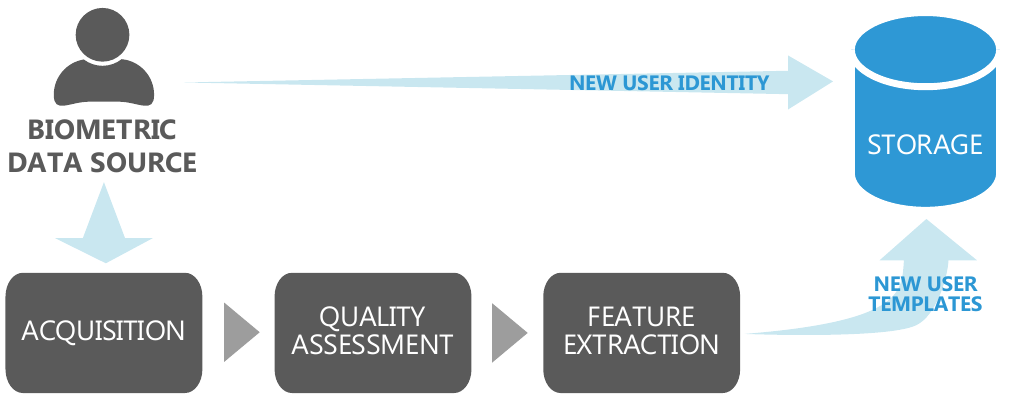}
       \caption{Enrollment}
       \label{verification}
  \end{subfigure}\\\vspace{0.03\linewidth}
  \begin{subfigure}[b]{0.45\linewidth}
       \includegraphics[width = \linewidth]{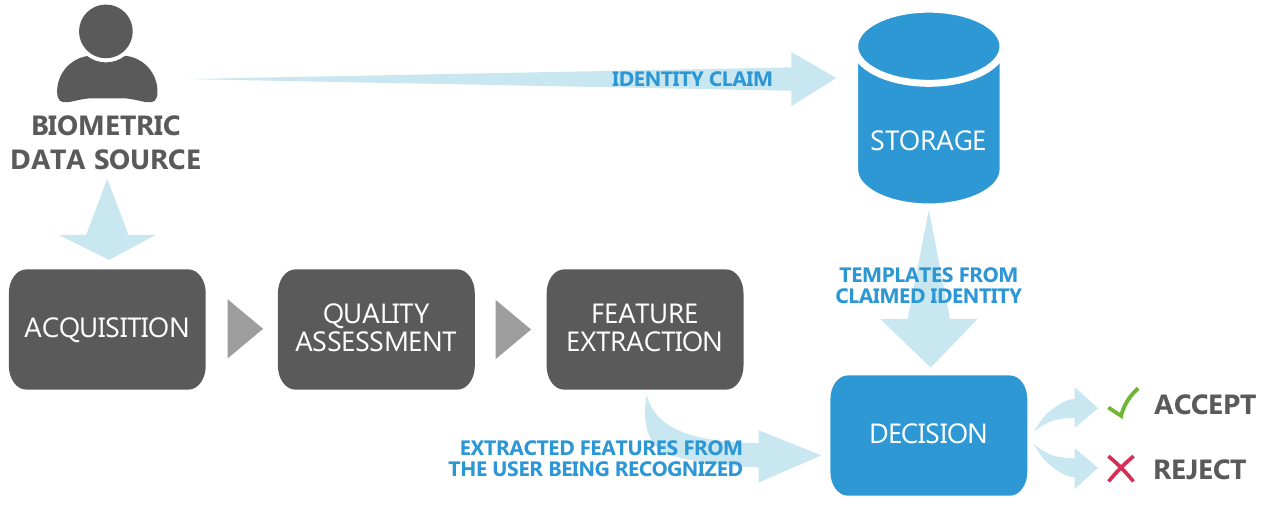}
       \caption{Authentication}
       \label{auth}
  \end{subfigure}\hspace{0.05\linewidth}
  \begin{subfigure}[b]{0.45\linewidth}
       \includegraphics[width = \linewidth]{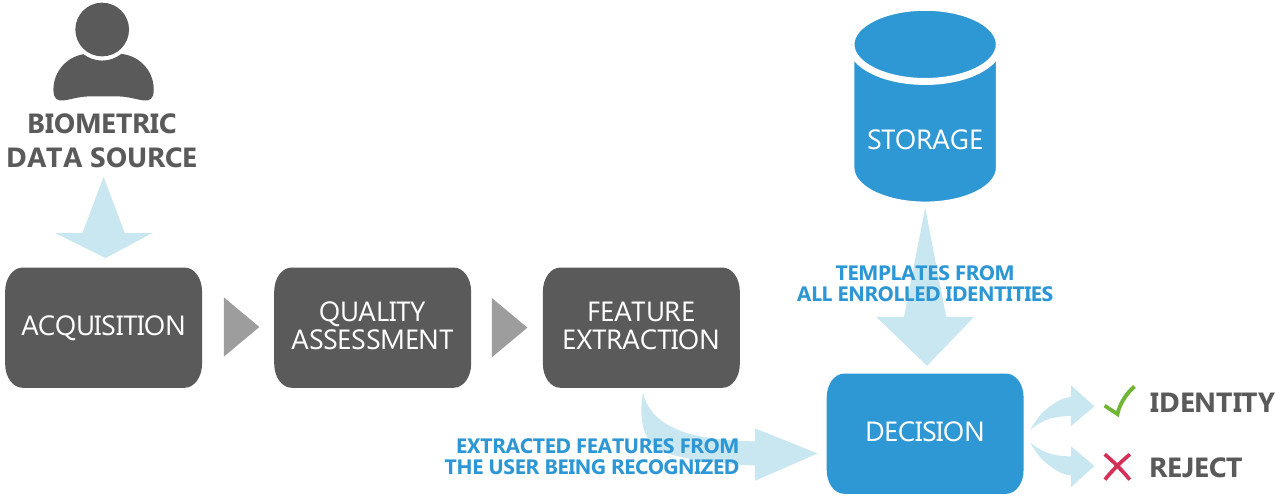}
       \caption{Identification}
       \label{identification}
  \end{subfigure}
  
  \caption{Different tasks of biometric systems. In (a) users add their information to the database. In (b) users claim to be someone and and the system attempts an 1:1 comparison. In (c) the system tries to match the user with all the identities in an 1:N problem.}
  \label{fig:modules_} 
  
\end{figure}

The word biometrics comes from Morris' definition in 1875 as the combination of the words Bio (life) and Metron (a measure). In other words, it consists of the usage of body traits and behavioural cues to perform measurement and analysis. This definition has endured over the years, and its applications evolved throughout the years. These systems are comprised of three major tasks: enrolment, authentication and identification (Fig.~\ref{fig:modules_}). Enrolment allows users to insert their biometric data into a gallery and link it to their identity. Authentication receives an identity claim and biometric information from the user, and performs a 1:1 verification with the provided information and the data in the gallery linked to the claimed identity. Finally, for identification, the system receives only the biometric data and tries to compare it with the entire gallery (1:N comparisons); if it does not match any identity it returns no identity. Within these tasks, there are secondary tasks that aim to secure the system, such as presentation attack detection and morphing attack detection. As mentioned before, the widespread usage of biometrics implies that all these systems are consistently identifying and authenticating humans.

As seen in Fig.~\ref{fig:modules_},  the majority of the modules are common to all the three main tasks. This structure redistributes responsibilities and allows for smaller and incremental improvements on specific modules. Their responsibility within the system is the following:

\begin{itemize}
    \item \textit{User}: Without the user, there is no biometric system. It is the most central role in the entire system, it provides all the data, and systems are designed to be used by them. Due to the strong regulations imposed by the European General Data Protection Regulation (GDPR)~\cite{EuropeanUnionGDPR}, the user is, more than ever, the central part of these systems.
    \item \textit{Acquisition}: The acquisition module is the bridge between the user data and the algorithm. Hence, as a bridge, it must comprise two different properties. First, it must be practical and benefit from a high usability score. Secondly, it must provide quality biometric data to the system and to the following modules. The acquisition process must be responsible for reducing the excessive noise and artefacts, which might require specialised hardware to read and digitise the biometric information.
    \item \textit{Quality assessment}: Despite the efforts applied to the acquisition module, some noise might still be present in the data, which might be caused by an intentional attack on the system. This module aims to further remove the noise, and discard attacks or unusable samples.
    \item \textit{Feature extraction}: Pattern recognition and signal processing based algorithms are capable of extracting meaningful features from input data. These extracted features represent the input signal mapped into a latent space that allows us to directly compare different samples. Researchers aim to create latent spaces that have a high variance between samples from different identities, and a low variance between samples of the same identity. In other words, discrimination between different identities is facilitated by this representation.
    \item \textit{Storage}: The storage of biometric templates is also an important element of biometric systems. On one hand, when the number of stored templates increases, the query response time and processing time are expected to increase too, and thus, this module is crucial for scalability. Moreover, there are security concerns and procedures that surround the storage of biometric templates, their link to the user identity and how they can be explored by harmful entities.
    \item \textit{Decision}: This final model is responsible for the decision. This is one of the most application-dependent modules, since, different threshold decisions and different application goals are involved. It can either return the Boolean result of an identity claim (authentication) or an identity (identification). Some systems further include a confidence metric that supports the usage of a reject-option for uncertain samples. 
\end{itemize}

Since the early 2010's these biometric systems have also benefited from the exponential growth of machine learning (ML) methods. ML-based systems excel in most of the artificial intelligence (AI) fields, and have outperformed other methods, as well as humans, in certain tasks. Undoubtedly, the success of AI systems is mainly due to three factors: i) improvements of deep learning (DL) methods; ii) availability of large databases; and iii) computational gains obtained with powerful graphics processing unit (GPU) cards~\cite{lecun2015deep, karpathy2014large}. DL-based neural networks extract feature representations from data, which are characterised by high compactness and strong discriminability.
In other words, the most affected module by the rise of deep learning systems has been the feature extraction module (Fig~\ref{fig:modules_}). It evolved from extracting handcrafted features to directly feed the inputs to deep neural networks. Feature selection is also optimised in a end-to-end fashion through the minimisation of the loss function. Moreover, the quality assessment module becomes less critical to the system since the features became significantly more robust to noise (for instance, models that can perform face recognition despite occlusions~\cite{neto2022beyond}). These performance gains were propelled by a significant increase in model complexity. In other words, the literature suggests that the opaqueness of a machine learning model may be positively correlated to its performance~\cite{rudin2019stop}. As seen in recent competitions dedicated to challenges related to face biometrics, the proposed solutions are mainly focused on deep learning approaches~\cite{boutros2021mfr,neto2022ocfr,huber2022morphing}. 

The existence of opaque models creates space for hidden biases, privacy issues, vulnerabilities to adversarial attacks and lack of transparency. And some of these vulnerabilities have already been shown to be present in some biometric systems~\cite{vakhshiteh2021adversarial,terhorst2021comprehensive}. In fact, it is challenging to directly detect them due to the complexity of the models used. Hence, the focus is being redirected into a) using models that are inherently interpretable; b) developing methods to explain the predictions of black box systems; c) incorporating fairness and other similar metrics into the optimisation process of the model. The relevance of these topics and research directions creates the need for a study that gathers the findings, successes, and most relevant research questions for explainable artificial intelligence (xAI) within biometric systems.

%Nonetheless previous successes and future directions of explainable artificial intelligence within biometric systems is not compiled nor it is easily available for researchers. 

%\hl{Focus of the paper}

The main contributions of this paper includes a chronological road-map of the shift of biometric systems into black box ones, the presentation of the current challenges of these systems,  the description of the available methods for explainable artificial intelligence, and their limitations. Concluding with the anticipated future directions of xAI applied to biometric systems. It includes a review of why explainability is necessary, an analysis of works both within biometric and computer vision domains, and future directions. Hence, besides this introductory Section, the document is divided into four other major Sections. First, Section~\ref{sec:two} discusses the current state of biometric systems as well as the need to integrate explainable artificial intelligence methods in their design. Afterwards, fundamental concepts and the background of xAI approaches are described in detail in Section~\ref{sec:three}, which finalises with methods that already started to merge xAI and biometrics. This Section is followed by Section~\ref{sec:four} that aims to illustrate the benefits of this merge and potential strategies to integrate both research areas. Finally, Section~\ref{sec:five} concludes the main ideas of the paper and presents future research directions that aim to develop fair and transparent biometrics systems.

\section{Why biometrics is not interpretable (but should be)}
\label{sec:two}
Before delving into explainable artificial intelligence concepts, it is important to understand how biometric systems evolved through time. Hence, this Section presents a brief introduction to the reasons that led biometric systems to become black box and the reasons that led researchers to focus on increasing the understanding of the inner workings of these systems.

% %-------------------------------------------------------------------------
%\subsection{A historical context of biometrics }
\subsection{How did biometric systems become opaque?}
% %\textcolor{magenta}{AFS}
% %\textcolor{blue}{Topic: i)how biometrics has conquered every aspect of our life; ii)The evolution of computational power and the rise of deep learning;}

Biometric systems are everywhere, from smartphones and laptops to border control and other sensitive areas. Both experts and non-experts have constant contact with these systems, but do they really understand what is under the hood? Do they trust biometric systems? To better understand the deep implications of these questions it is necessary to understand the current state of biometrics systems and their evolution over time. 

Biometric systems have been used since late 1800's, with the introduction of fingerprints in 1892~\cite{galton1892finger,ellenbogen2012reasoned}. After a few years their underlining methodologies have started to follow the automation path since the mid-1900s. From these initial applications to the current methods, there has been significant progress regarding the precision and accuracy of the overall system. More concretely, the first known biometric system to benefit from automation was fingerprint-based biometrics, which started with methodologies focusing on minutiae points~\cite{jain1997line}, replicating the human analysis. Then the methodologies moved to textural approaches, and currently, the DL-based techniques are predominant for this biometric trait~\cite{minaee2023biometrics}. Face biometrics started as a technique based on specific regions of the face - fiducial points - and was the first biometric trait to truly leverage DL methods~\cite{taigman2014deepface}. Iris recognition was tackled with texture-based methods due to the texture richness of these images. Afterwards, this trait has also moved towards DL approaches~\cite{minaee2016experimental}.

Over the previous decade, researchers, excited with the quick progress, focused on improving the performance of deep learning-based biometric systems. As reported in the Face Recognition Vendor Test conducted by NIST~\cite{grother2018ongoing}, the massive gains in face recognition accuracy have coincided with the usage of convolutional neural networks to extract an array of features from a face image. It is also worth noting how significant these improvements (from 2013 to 2018) are when compared the ones made during the previous period (from 2010 to 2013)~\cite{grother2018ongoing}. For instance, in a four-year period the performance of the best performing face recognition algorithm improved $\sim$20x~\cite{grother2014face,grother2018ongoing}. From the initial False Negative Identification Rate (FNIR) of 0.041, achieved by EC0C in 2014, to a FNIR of 0.002, attained by NEC-3 in 2018. In practice, this translates into a $\sim$20x smaller error. The rapid improvements of these models correlated positively with their opaqueness. And, despite their remarkable performances, these systems are often deployed in use-cases with serious implications regarding wrong predictions. Moreover, discrimination and other harms might cause these errors to impact more a certain demographic group. As such, fairness and transparency must be tackled together with the development of better, more efficient and accurate systems.

%%\textcolor{red}{SHOULD WE FURTHER DEVELOP THIS SUBSECTION? EVEN MORE DETAILS? SEEMS A BIT SHORT, I AM NOT SURE.}
%%\\
%%\textcolor{olive}{sim...}
%%% DIDN't have time. Maybe for the final version

\subsection{Why Biometric systems need to reclaim transparency?}

Following the idea of the previous section, explainability should become a central piece in the development of biometric systems. The predictions given by a biometric system should be sustained by an explanation or an interpretable behaviour.  Yet, the transparency and the interpretability of these systems are, as of now, negatively correlated with their performance~\cite{guidotti2018survey}. DL tools can be a double-edged sword. On the one hand, it provided grounds for achieving human-level performance, on the other hand, it was a step away from white-box approaches. The complexity and capability to fit the data displayed by biometric systems is also a potential cause for their sensitivity to the underlying demographic characteristics of the training data, resulting in performances that vary with the demographic attributes of the user. For instance, this discriminatory behaviour can be harmful if a human is not aware of this behaviour, and it can be displayed through the usage of non-causal attributes or lack of discriminative features for that specific demographic group. Wang~\textit{et al.}~\cite{wang2021meta} have shown that the usage of a shared optimisation process for face recognition can penalise underrepresented, or more difficult skin tones. Similarly, Hull~\cite{hull2023dirty} discussed the implications of spurious correlations of dirty data on the models' predictions. It is important to retain that, differently from other fields, on biometrics, sensitive information such as gender, age, ethnicity and health status can be implicitly encoded in the inputs~\cite{mirabet2023new,mirabet2023lvt,mirabet2022deep}, or in the extracted features~\cite{terhorst2020beyond} and are not easily removable. For such reason, these systems and their data require several layers of protection and privacy. Neto~\textit{et al.}~\cite{neto2023compressed} have shown that a simple classifier is able to learn how to separate different ethnicity groups on the latent space of face recognition deep neural networks. 

\begin{figure}[!h]
\centering
\includegraphics[width=4.8in]{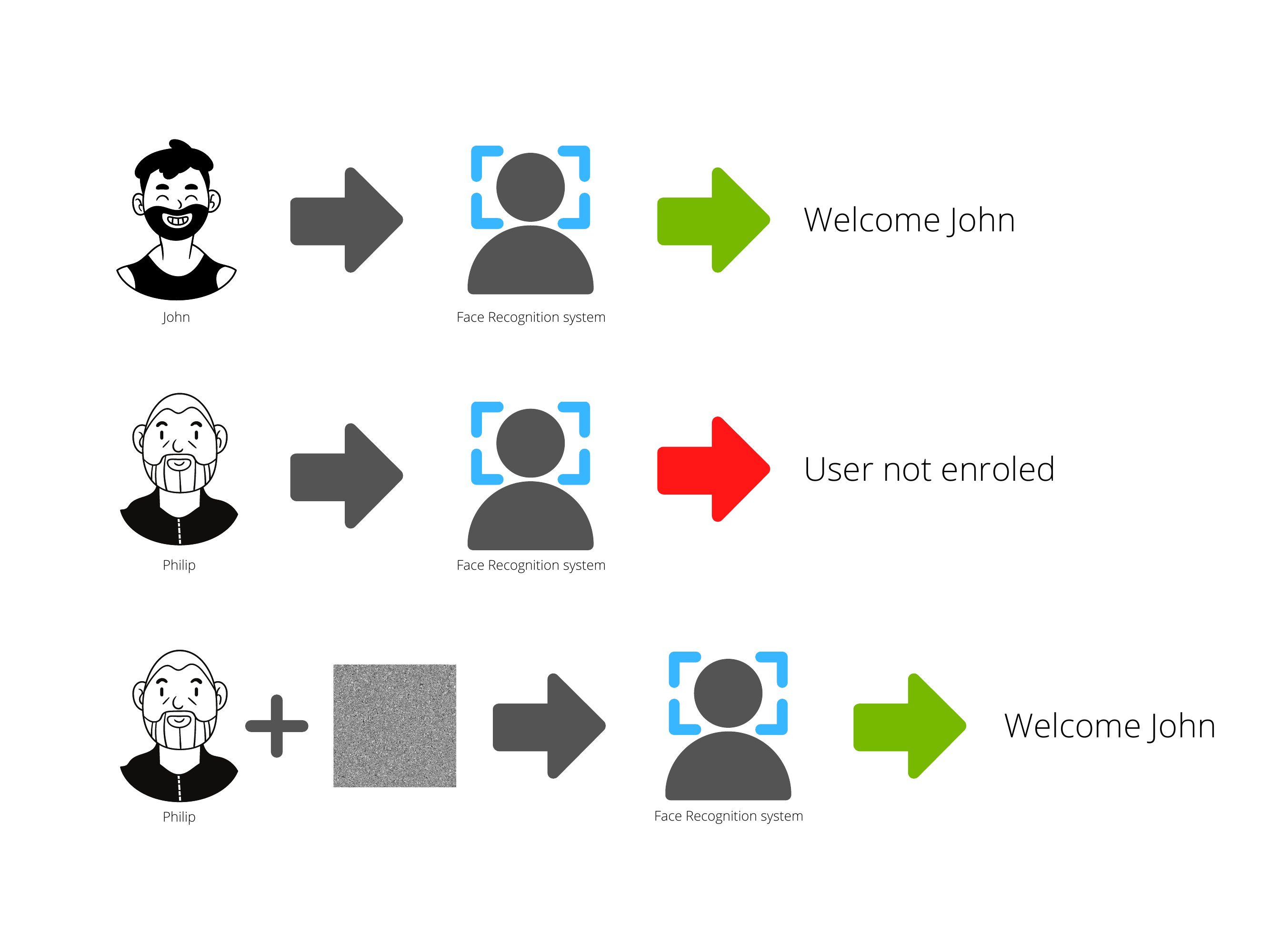}
\caption{Vulnerability of face recognition systems to natural-looking adversarial input images that drive the model to incorrect output predictions.}
\label{fig:adversarial}
\end{figure}

Their black-box nature provides a grounding for targeted attacks. For instance, some adversarial attacks are capable of completely fooling a system by adding small noisy patterns to the image without affecting its realism to humans~\cite{Dong_2019_CVPR,zhong2020towards} (Fig.~\ref{fig:adversarial}). Other attacks consist of including patterns, for instance on the outfit of a person, that make that person undetectable~\cite{yang2020design,xu2020adversarial,thys2019fooling}. Moreover, research can be found that focus on backdoor attacks, which encode behaviours hidden within the network and can be triggered by a certain input~\cite{xue2021backdoors,unnervik2022anomaly}. These attacks can be difficult to detect, especially if we are not aware of the expected behaviour of the system. Besides these attacks, which are common within the deep learning literature, biometrics systems are targeted by several other attacks, such as: presentation~\cite{yambay2019review,yambay2023review}, morphing~\cite{huber2022syn} and deepfake attacks~\cite{peng2021dfgc}. Recently, beautification filters, similar to the ones found on  popular social networks, have also shown a potential to impact these biometrics systems~\cite{mirabet2022impact}.

For better performances, researchers are also trying to create multi-modal systems based on more than one biometric trait~\cite{bala2022multimodal}. The resulting model is expected to be more robust, but at the same time more complex. Considering the already difficult task of explaining and interpreting a model designed to work with a single modality, it might be significantly harder to understand a model with multiple traits. If one of the traits can be explained, then it might lead to improvements in the overall explainability of the remaining traits~\cite{pinto2023seamless}.

Opaque systems are not trustable~\cite{von2021transparency,duran2021afraid}, sometimes not even if an attempted explanation is provided~\cite{lakkaraju2020fool}. In other words, black-box systems reduce users' trust, especially after a wrong prediction, since it hinders the ability to follow the \textit{thought process} of the system. Due to its prominence and effects on users' trust, the use of black-box systems becomes one of the most impactful factors for the adoption of new technologies. Educating users' in the recent advances of biometric technology is crucial but fails to compensate for the lack of explainability of these systems. As such, the adoption is conditioned on the ability to transpose the thought process of the model to a common language between the system and the user. Moreover, due to the heavy use of personal data, users must be protected through regulations and inspections. A strong protection, especially if perceived by the users, further increases the trust in the system~\cite{miltgen2013determinants,abdulkareem2023evaluating}. With this in mind, several countries and organisations worldwide have defined some degree of regulation of biometrics. For instance, the European Union (EU), with their General Data Protection Regulation (GDPR)~\cite{EuropeanUnionGDPR}, limits the processing of biometric information. Exceptions include law-related scenarios and if the user explicitly consents. In the United States of America (U.S.A.), at least three different states (Illinois, Texas and Washington) have laws heavily regulating biometric systems~\footnote{\url{https://www.huschblackwell.com/2023-state-biometric-privacy-law-tracker}}.

For years, researchers and engineers have joined forces to deploy several biometric systems at scale in real world scenarios. Despite their attempts to promote trust in those systems, and to prove that the systems have an unbiased and accurate behaviour, they have been shown to be wrong ~\cite{Drozdowski2020,Cook2019,Srinivas2019,Klare2012,Kortylewski2018} and this technology has been described as not being production-ready. Srinivas~\textit{et al.}~\cite{Srinivas2019} have shown a performance difference between adults and children. Klare~\textit{et al.}~\cite{Klare2012} added both ethnicity and gender to the age as factors with an impact on the performance of face recognition systems. These consecutive and recurrent mistakes undermined the trustworthiness of these systems. In other words, the deployment of biased systems puts at risk the affected minorities, and further hurts the field as well as its growth potential. The first step for a better adoption of biometric systems is undoubtedly to provide users with an explanation of the algorithm decisions~\cite{dhar2023challenges,das2020opportunities}.

The interpretation of black box models and their conversion to white systems has several purposes. Not only can they be used to provide explanations and answers that incite the user to believe in the answer given, but they can also be used to detect unexpected behaviour, attacks or biases. For this reason, developing interpretable models or explanation methods is crucial for the smooth implementation of biometric systems in high-risk situations. The relation between risk, trust and bias has been studied before by Lai \textit{et al.}~\cite{Lai2020} and it highlights the need for transparent and explainable systems in order to achieve safe, trustworthy and unbiased models. With this in mind, researchers can aim to a careful deployment of biometric systems in large scale, further improving their capabilities and designing software that can work together with humans for more accurate solutions.

\section{Fundamental concepts and background}
\label{sec:three}

As with any recent scientific field, the terminology of explainable artificial intelligence is surrounded by fog. There are several interpretations across the literature of the same terms.  Hopefully, over time, the meaning of each term  will converge to common ground. %However, this article intends to be a reference for xAI applied to biometrics. 
The first part of this Section is dedicated to defining these terms as they are used in the literature and to shed some light on how their definition can be improved. Once established the proposed interpretation of xAI terms, the second part of this Section discusses interpretability and explainability re-framed to a novel interpretation of these terms through the lens of causality. Finally, the third and last part of this Section presents existing biometric works that already include xAI concepts.
%----------------------------------
\subsection{Towards a common terminology for xAI in Biometrics: Can we speak the same language?}
\label{subsection:terminology}
%%\textcolor{blue}{Topic:(terminology to be used throughout the paper) }

%%\textcolor{magenta}{AFS}
%\textcolor{yellow}{Topic: too many available; provide a “working definition”}

The definitions of interpretability and explainability found in the literature are not consistent between different authors. Some authors use them interchangeably, while others make a distinction between them~\cite{montavon2018}. Within the group of researchers that make a distinction, the understanding of these terms also varies. Hence, we  aim at establishing a working definition to be used in the following Sections of this document. 

In our definition, explainability and interpretability are different but not independent concepts. Whose definition is linked to two main questions: ``How?'' and ``Why?''. The former is related to interpretability, which aims at providing a clear understanding of the inner workings of a model and the reasoning process that leads to an output. The latter is the question that, given a model's prediction, an explanation tries to answer~\cite{gilpin2018explaining}. However, it is possible to deduce the second question from the first, because if we fully understand the behaviour of a model, we also know the answer to the ``Why?'' question. The opposite is not true, because knowing the explanation of a certain prediction does not directly indicate how the model reached that prediction. We further introduce a third question, "Where?", which is frequently associated with explainability, while not providing concrete answers to the "Why?" question. 

Due to the ability to answer to the ``Why?'' question, by answering the ``How?'' question, it can be assumed that knowing the answer to the first, it might be the case that we are moving towards the answer to the second, even if we do not have enough information to answer it completely. We believe that there is a strong and close relationship between xAI and causality~\cite{pearl2009causality}. 

\begin{figure}[!t]
    \centering
    \includegraphics[width=\linewidth]{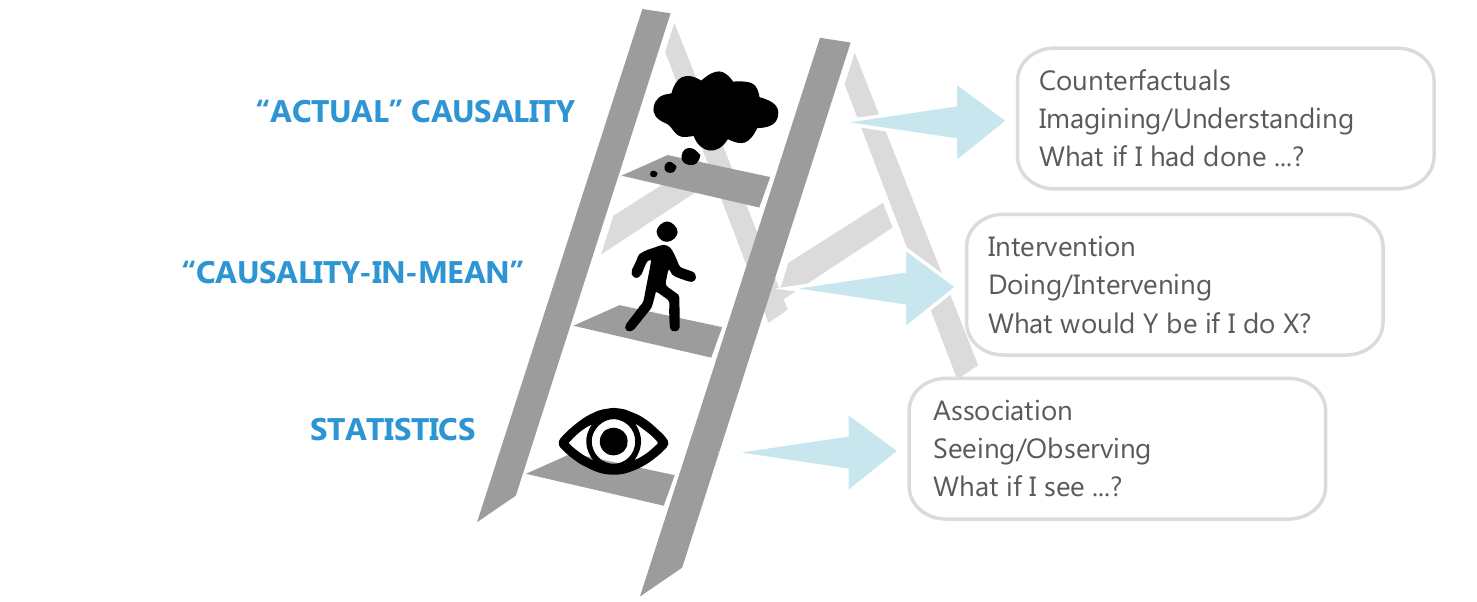}
    \caption{The Ladder of Causality (adapted from~\cite{pearl2018book}).}
    \label{fig:causality_ladder}
\end{figure}

%\textcolor{olive}{afs: 10nov2023} 

Causality studies the relationship between causes and effects, and it aims to model the causal relationships that define an event or sequence of events~\cite{bunge2017causality}. Similarly, explainable artificial intelligence aims to understand the model inner mechanisms that led to a certain output when conditioned on a certain input. Pearl~\cite{pearl2009causality,pearl2018book} proposed the ladder of causality (Fig.~\ref{fig:causality_ladder}). Within this framework, it is possible to characterise causality in three levels: (1) association, (2) intervention and (3) counterfactuals. Each of these levels aims to answer a certain type of questions, and they follow a hierarchical structure, in the sense that for answering the questions of a level $i$, we must have the information to answer the questions of all previous levels~\cite{pearl2009causality}.  Pearl~\cite{pearl2009causality} goes beyond defining the degree of causality associated with each level, he further establishes the mathematical notation for each level. For the Association level (1) the usual representation is a conditional probability in the form of $P(y|x)=p$, which calculates the probability $p$ of the event $Y=y$ given that the event $X=x$ was observed. Bayesian networks are effective and efficient at calculating these conditional probabilities~\cite{pearl1985bayesian}. Intervention level (2) operations can be described, leveraging the \textit{do} operator, as $P(y|do(x),z)$, which calculates the probability of the event $Y=y$, given that we intervene to set $X$ to $x$, and subsequently observed the event $Z=z$.  The third and last level, the Counterfactuals, can be represented by $P(y_x|x',y')$, which states the probability $P(y|x)$ when we actually observed the events $X=x'$ and $Y=y'$.  It is only possible to compute these if one has access to functional or Structural Equation models, or their properties. 

The connection between these two scientific fields can be used to re-frame xAI, and we suggest this change in the following sections.   

%\textcolor{blue}{(... provide terminology to be used throughout the paper) }

%%---%%
%\subsubsection{ Historical context}
%\textcolor{yellow}{Topic: (example: why decision trees appeared?)... }

%%---%%
\subsection{A brief introduction to xAI}

Some of the terms used to define explainable artificial intelligence methods have overlapping definitions and concepts. Some authors aim at dividing them into pre-, in-, and post-model~\cite{kim2017interpretable}, others in methods specific and agnostic to the deep model, and even categorizing them as post hoc or not.

By dividing the interpretability into three smaller subtopics, researchers allow for the creation of different goals for each task. The pre-model analysis focuses on data knowledge and understanding. It significantly leverages visualization techniques that allow researchers to understand the distributions of the input samples~\cite{varshney2012interactive}. This is an approach frequently used in Business Intelligence (BI) use cases. Traditionally, this technique led to the improved capability of creating handcrafted features for simpler models; nowadays, it is quite relevant to detect some biases in the data, which can be described through the analysis of the skewness in the dataset~\cite{wu2023real}. Understanding the data used for training allows for increased confidence in the posterior decisions and explanations. 

\begin{figure}[!t]
    \centering
    \includegraphics[width=\linewidth]{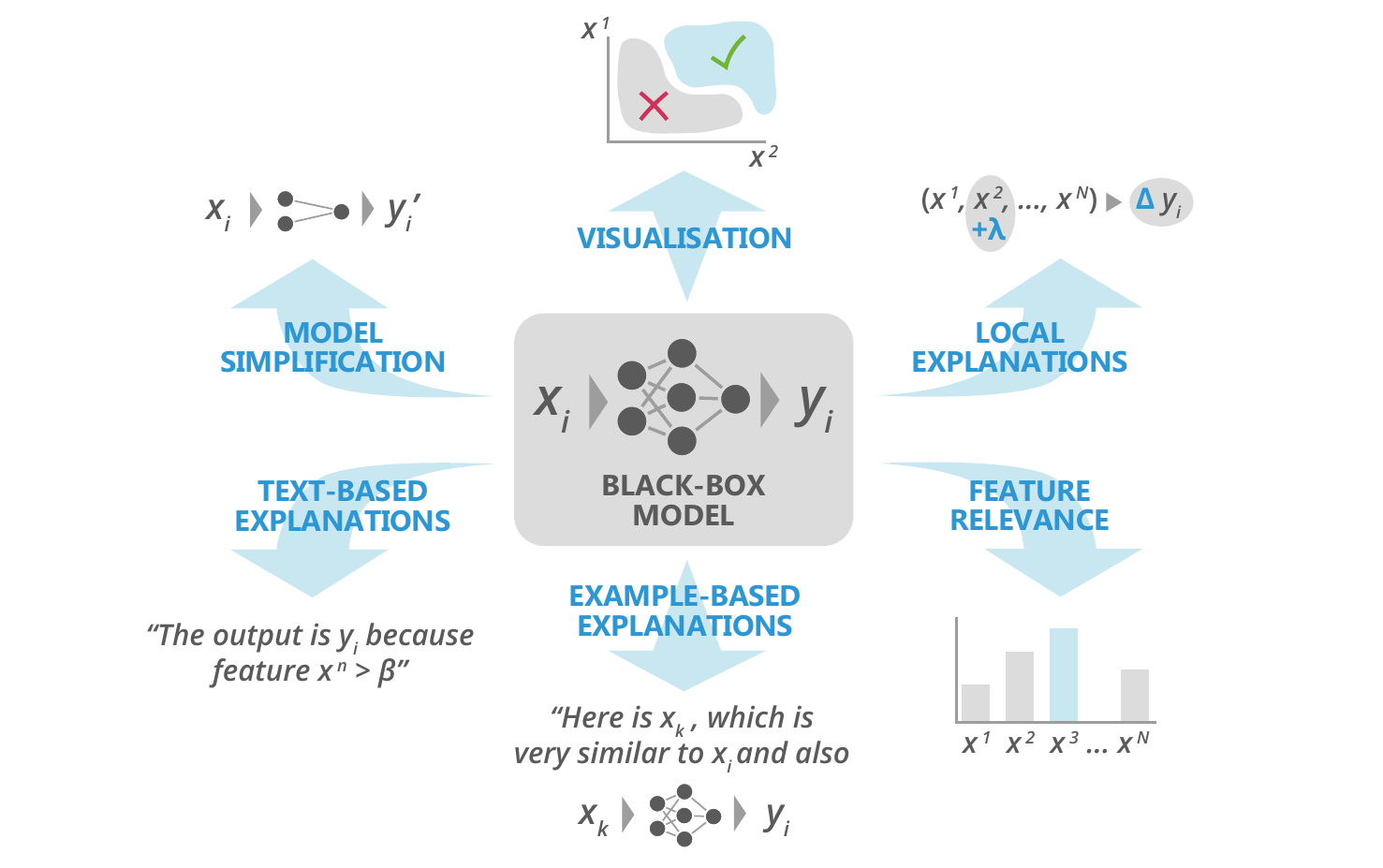}
    \caption{How we can explain black box models: a summary of current \emph{post hoc} explainability approaches (adapted from~\cite{BarredoArrieta2020}).}
    \label{fig:diverse_forms_explanations}
\end{figure}

On the other hand, there are also techniques that can be applied to the model and its outputs, during and after training. When the training process is conditioned by some constraint, these are known as in-model techniques and they focus on the direct integration of interpretability into the model through the usage of these conditioning priors. Some families of machine learning models already include inherently/intrinsically interpretable mechanisms in their design.  Hence, their use is also considered to be an in-model approach.  Some of these interpretable models are rule based~\cite{breiman1984classification,rivest1987learning,wang2017bayesian},  per-feature based~\cite{hastie1986gam,fico2006scorecards} and some simple versions of linear models such as linear regressors. They are, however, limited by the semantic meaning of the original features and the size/complexity/depth of the model. Nonetheless, it is possible to increase, to a certain degree, the interpretability of a complex model. Constraints to these models, frequently expressed in the form of regularisation terms (such as L1 regularisation~\cite{goodfellow2016deep}), are the most common approach to increase the interpretability of those models. Other approaches include constraining the model to have certain properties, such as monotonicity~\cite{sill1998monotonic}. These constraints can, sometimes, be applied or enforced in specific modules. While, it is known that achieving global holistic interpretability is an extremely difficult task, analysing the model in modules for global modular interpretability is much more manageable~\cite{molnar2019}. The exploration of these in-model approaches within the context of biometric systems is rather limited due to the potential sacrifice of predictive power. Instead, the focus has been on \emph{post hoc} methods, which is a different term to post-model approaches. It is, however, important to note that, if the constraints applied to the model represent meaningful priors, it can be the case that a better solution can be found, since the search space is reduced to exclude worse solutions~\cite{Neto2022OrthMAD,Xin22rashmon}.

Aligned with the efforts towards continuing the development of more accurate models, the focus has been partially redirected towards post-model interpretability. This approach is frequently known as explainability since these methods generate explanations to a given prediction of a trained model. While it is more common to have model-specific in-model approaches and model-agnostic post-model approaches, this not always the case. For instance, gradient-based methods, which leverage gradient information to identify areas of the image that contribute to the final decision, are specific to deep neural networks~\cite{baehrens2010explain,simonyan2014deep,springenberg2014striving,smilkov2017smoothgrad}. One of the common approaches to post hoc interpretability adds perturbations to the input in order to assess the impact on the output~\cite{fernandes2018deep}. As discussed in the following Sections, these methods do not have a rich semantic meaning. Hence, to enrich the explanations produced, some methods map the latent features of a model to the input space~\cite{zeiler2014visualizing,dosovitskiy2016inverting} or to a representation of semantic concepts~\cite{bau2017network}. 

Lately, there has been research on a post-model approach that closely resembles ideas from in-model methods. While training a simpler model is somewhat complicated if the task is too complex, it is possible to teach a simpler model to mimic the behaviour of the more complex (deeper) model~\cite{ba2014do}. Due to the lower parameter count, simpler models present a more interpretable view of the decisions made by the complex model. This is known as model distillation. This and other compression techniques, such as compression~\cite{boutros2022quantface} and pruning~\cite{lin2022fairgrape} can help to reduce the complexity of an overly complex neural network architecture~\cite{caldeira2024model}.

In biometrics, the prevalence of pre- and in-model approaches is rather limited when compared to post-model methods. Due to the high requirements for accuracy and other performance metrics, the usage of post hoc methods is much more convenient, requires less engineering effort, and provides an acceptable trade-off between accuracy and these efforts. Nonetheless, while it provides immediate convenience, it is not closer to a proper path to xAI than the other two approaches. Moreover, the current taxonomy is limited as it does not explore the real meaning of understanding a specific model. Rather, current approaches focus on explanation methods that focus on understanding the prediction, but not the model. Furthermore, this taxonomy fails to include several in-model approaches that improve the interpretability of the system through the introduction of domain knowledge within the training process. As such, the following Section proposes a revised taxonomy for Explainable AI and how it encapsulates the methods seen in Figure~\ref{fig:diverse_forms_explanations}.

\subsection{The causal relationships of a deep model}

Causality's main goal is to understand and learn the causal model of the \textit{world}. Once this model is found, it is possible to make predictions and informed guesses with a tremendous degree of certainty. It is possible to grasp the implication of the smallest details on the deviations of a certain event. As previously mentioned, the ladder of causality represents a simplification of the steps one must take towards getting a causal model of the world. In practice, the last step on the ladder represents a mental representation of the \textit{world} that allows humans to replay, recreate and even imagine new events on their heads. One can formulate the question of "What would happen if I stop taking my antibiotics now?", and humans are capable of answering through imagination. 

\begin{figure}[h!]
    \centering
    \includegraphics[width=\linewidth]{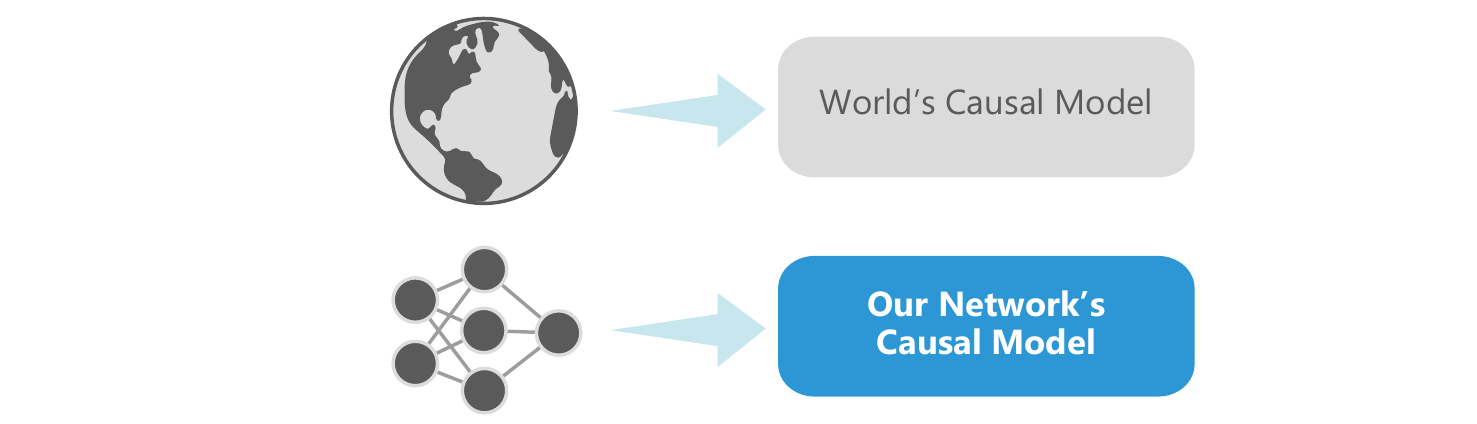}
    \caption{A figure illustrating the World's Causal Model which represents all the causal relationships that we observe and rule the functioning of the world. Below we see a representation of the Causal Model of a Deep Neural Network. While the latter represents the causal relationships within the "rules" learnt by the DNN, the former represents the true causal relationships of the world. These two models are very likely to disagree, and the latter does not aim to explain the causal relationships of the world, just the specific deep learning model it represents. }
    \label{fig:worldcausalmodel}
\end{figure}

Explainable artificial intelligence shares the same goal as causality approaches, in the sense that it aims to understand the model and provide explanations for the relation between two or more events (e.g. input and prediction). This means that, in a sense, we can try to frame xAI within the context of causality. And, for that, it is necessary to answer the following question "Which causal relationships should our causal model learn?". Since we are only trying to understand a specific model, our causal system does not need to represent the \textit{world}. Instead, we want to find a causal system that models the causal relations of our deep model, as show in Figure~\ref{fig:worldcausalmodel}.

When compared to the previous approach, there are significantly more advantages to this approach. First, it is possible to measure the transition between the different steps of causality. It is also possible to quickly detect spurious connections learnt by our model, through a comparison between its causal model with reality. For instance, if a model uses the skin tone for a task that should not use that specific feature, the causal model of the deep model shall provide this information. Hence, hidden biases can be detected and explained in a straightforward manner. In other words, we can look for misalignments between the causal relationships of \textit{reality} and the causal relationships of a deep model. So, in a biased model, we would find, for instance, a causal relationship between gender and lower wages, and through a careful comparison with the \textit{world} causal relationships, we can deem this specific relationship as biased. %Finally, if causal model of our model is known, it can also be used to produce \textit{"imaginative"} outputs in the form of a counterfactual. }

\begin{figure}[h!]
    \centering
    \includegraphics[width=\linewidth]{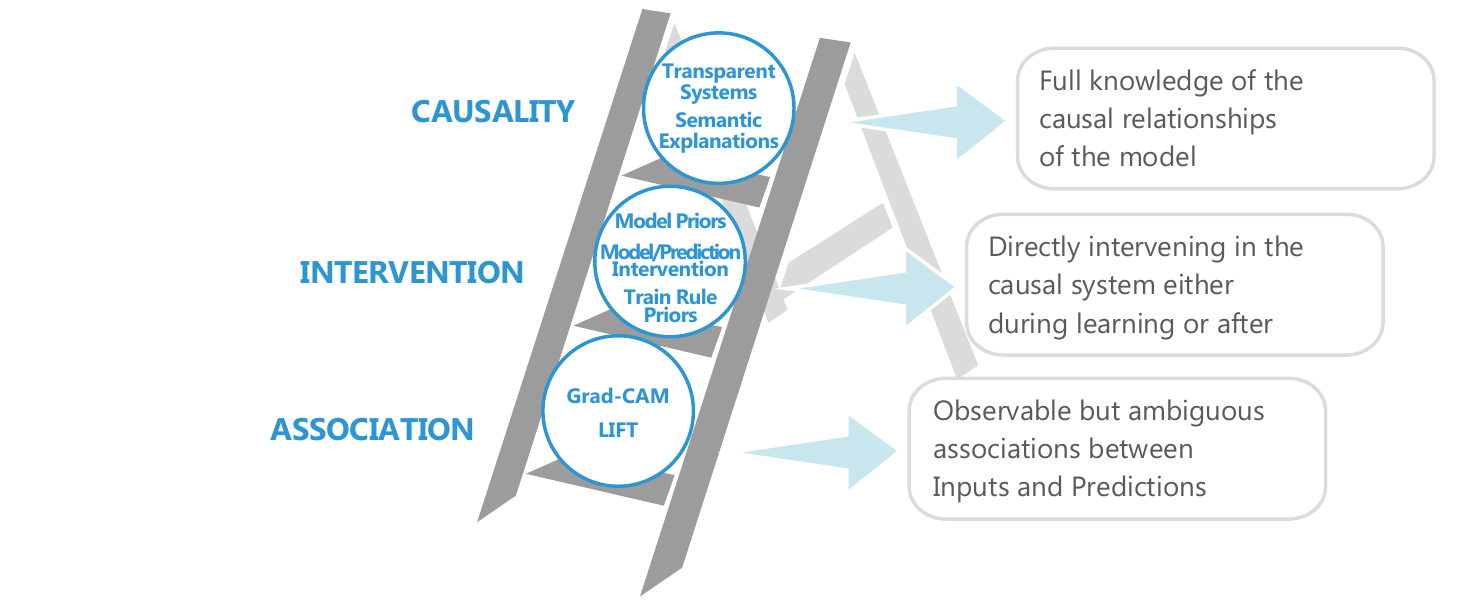}
    \caption{xAI Ladder: causality ladder repurposed to categorise different types of xAI approaches according to the understanding of the model provided by them.}
    \label{fig:ladderxai}
\end{figure}

For this we revisited the ladder of causality (Fig. \ref{fig:causality_ladder}) and propose a similar structure for the ladder of explainable artificial intelligence as seen in Figure~\ref{fig:ladderxai}. This structure for categorising different xAI approaches shows how they contribute to answering the questions that we introduced in Section~\ref{subsection:terminology}. Furthermore, it highlights the importance of each technique and the journey ahead. On the first step of the ladder, we can see observational explanations, such as visualisation maps, which are quite powerful at locating elements that were relevant for the decision, but that do not hold semantic value or any information regarding what was in fact important for the decision. This presents itself closely linked with the associations at the first step of the ladder of causality. Similarly, pre-model approaches, that rely on the understanding of the data, are unaware of the inner mechanisms of the model, hence, these approaches cannot, by definition, go beyond the first step on the ladder. 

The second step on the xAI ladder encompasses some of the more advanced approaches currently used in the literature. Furthermore, it is the case that an approach might be between two different rungs. For instance, one might insert prior causal knowledge to condition the training process. Thus, it is possible to know the inner workings of the model regarding that prior, which means that it is possible to reduce the set of possible explanations while partially knowing a subset of the causal relationships of our deep network. As such, the rung climbing is not a discrete process, but instead a continuous one. Some of the most common methods on the second rung are case-based explanations, semantic explanations, LIME (which closely resembles the $do(.)$ operation~\cite{pearl2009causality} by intervening in the input and analysing its effects. But in fact, the majority of applications is in between the first and second rung, where they can perfectly answer the "Where?" question regarding the importance of the variables, but cannot provide a strong and complete answer the "Why?" question. Climbing from the second to the third rung is the final goal of researchers working on xAI. This last rung requires a complete understanding of the inner causal relationships of the deep network. When this happens, one can fully answer the "How?" question, and can further answer to the "Why?" and "Where?" questions. Moreover, knowing these relationships means that producing counterfactuals is not a difficult task, for instance, for a skin-tone biased model we can quickly verify that a change in the skin tone would impact the final decision. And for this, we are not required to provide a new empirical sample to the model so we can empirically check the result (as happens with LIME for instance). 

The implications, nuances and adaptations of this new framework to categorise are vast, and require a careful analysis and continual study. As such, we provide in the following sections a few notes on each rung. 

\subsubsection{The First Rung: Observational methods as a first order explanation} 
% Pedro: Mudei o título da subsubsecção. Visualization versus explainability era muito mais "forte" 
% Tiago: Start

As with the association within the causal domain, in xAI, visualisation approaches are the most common. This tendency can be explained by the fact that most of the visualisation approaches require little computational effort, are less dependent on the model architecture and work well for several tasks. This new definition categorises visualisations as first order explanations, meaning that these are the most immediate explanations one can generate to quickly inspect the reasons behind a prediction. Furthermore, they provide less information than second and third order explanations. 

One of the first strategies for the visualization of deep learning models was proposed by Zeiler~\textit{et al.}~\cite{zeiler2014visualizing}. This was mainly motivated by the scarce insights that researchers had about the functioning and behavior of the models that led them to achieve high performances. In their approach, the authors used a multi-layered deconvolutional network~\cite{zeiler2011adaptive} to project the feature activations into the input space. Besides, they also proposed several occlusion tests to evaluate the sensitivity of the classifier. They occluded portions of the input image and revealed the feature activations, thus showing the parts of the inputs that were relevant to the classification.  Later, Simonyan~\textit{et al.}~\cite{simonyan2014deep} proposed two techniques that focus on the respective class of the input to generate the feature activations. This is achieved through the computation of the gradient of the class score concerning the input image. In the first use case, the authors built their work on top of Erhan~\textit{et al.}~\cite{erhan2009visualizing}. Given a score, and a learned model, the idea is to generate an image that maximizes the score, outputted by the model. This synthetic image is thus representative of the class appearance learned by the model~\cite{simonyan2014deep}. The second use case applies a similar strategy to generate a class~\textit{saliency map}. In this approach, the idea is to generate the saliency map that maximizes the class of a given input image. The result of this method is a mask that has pixels with high intensities in the locations that are related to that class. 

Springenberg~\textit{et al.}~\cite{springenberg2014striving} revisited the work behind deconvolution and performed experiments with a model composed solely of convolutional layers. Known as~\textit{guided backpropagation}, this approach proposed a modified way to handle backpropagation through rectified linear (ReLU) nonlinearity: instead of removing the negative values corresponding to negative entries of the top gradient (\textit{i.e.}, deconvolution) or bottom data (\textit{i.e.}, backpropagation), they remove the values for which at least one of these entries is negative. According to the authors, this technique prevents the backward flow of negative gradients, which often correspond to the neurons which decrease the activation of the higher layer unit one aims to visualize. Sundararajan~\textit{et al.}~\cite{sundararajan2017axiomatic} argued that most attribution methods did not respect two fundamental axioms (\textit{sensitivity} and~\textit{implementation invariance}), thus representing an important flaw of such methods. Taking this premise into account, the authors proposed a method called~\textit{integrated gradients}, which does not require any modification to the original network architecture, while being related only to the gradient operator. Following a similar research line, Shrikumar~\textit{et al.}~\cite{shrikumar2017learning} created DeepLIFT. This method is based on the backpropagation of the contributions of all the neurons in the network to all the features of the input and works by comparing the activation of each neuron to a given \textit{reference activation} and assigning contribution scores according to this difference. DeepLIFT also allows practitioners to give separate consideration to positive and negative contributions, hence allowing them to find other dependencies. 

Bach~\textit{et al.}~\cite{bach2015pixel} studied the influence of pixel-level contributions through the introduction of the~\textit{layer-wise relevance propagation} (LRP) technique which considers that a deep model can be divided into several layers of computation which may be responsible for extracting features or performing classification. Following the work of Zhou~\textit{et al.}~\cite{zhou2016learning} on~\textit{class activation mapping} (CAM), Selvaraju~\textit{et al.}~\cite{selvaraju2017grad} developed the~\textit{gradient-weighted class activation mapping} (Grad-CAM), which allows the practitioner to generate saliency maps with respect to a specific class (\textit{i.e.}, class-discriminative), a property that may help researchers unveiling some visual concepts related to a given class.

% Tiago: End

Frequently, these visualisations have not been seen as explanation methods, instead, they have been seen as a tool that can be used to provide additional information to the end-user~\cite{samek2017explainable,Chatzimparmpas2020}. In a sense, it can be seen as a parallel of distillation methods frequently applied to complex models, but in this case, used on complex high-order explanations. Even more than traditional explanations, the interpretation of outputs from a visualisation method is highly subjective and user-dependent, despite some attempts to objectively evaluate them~\cite{samek2016evaluating}. The former aligns with the perspective of assuming visualisations as approaches to achieve first order explainability, whereas the second cements it through another parallelism between visualisations and associations, which are also subject to very different interpretations. Hence, currently, saliency maps and other methods of feature visualisation are a bridge between the limited perception of humans on any number of dimensions above two or three, and the complete user-specific semantic explanations. 

The practicality of the visualisations as a tool for first order explanations led to a wide-spread implementation of this. And while researchers agree that using visualisation techniques has a positive effect on our understanding of the deep model, the amount of information contained within these explanations is rather limited and vague. For instance, we remove crucial information from these explanations, such as the importance of colour, transmitting a more open-ended problem to be discussed among humans. One of the most consensual statements between experts in explainable artificial intelligence is that humans can better understand counterfactual explanations~\cite{Van_Hoeck2015,Byrne2019}. In other words, a counter-example of the one given as input, which introduces a notion of contrastiveness~\cite{lipton_1990}. This behaviour is representative of the relationship between counterfactual examples and causality~\cite{Woodward2003,byrne_2007}. In other words, to move towards more complete explanations it is necessary to keep climbing the ladder.

Despite their limitations and their current use, it is expected that in the near future explanation methods will keep on leveraging methods from the first rung. Their practicality is difficult to defeat, and several tasks and problems are not concerned with higher-order explainability, since they only focus on small assessments of the model's behaviour.

It is worth noting that while this first rung considers both post-model approaches (e.g. Grad-CAM) and pre-model approaches (e.g. data visualisation), these can still be differentiated as some of them already have some information regarding the model's architecture and gradient propagation. Hence, it can be said that while not providing an answer to the "Why?" question, they provide stronger insights into the behaviour of the model. The following sub-section keeps on climbing the ladder.

\subsubsection{The Second Rung: Higher Order Explanation with Priors and Semantic Information}

\begin{figure}[h!]
    \centering
    \begin{minipage}{0.45\textwidth}
        \centering
        \includegraphics[width=\textwidth]{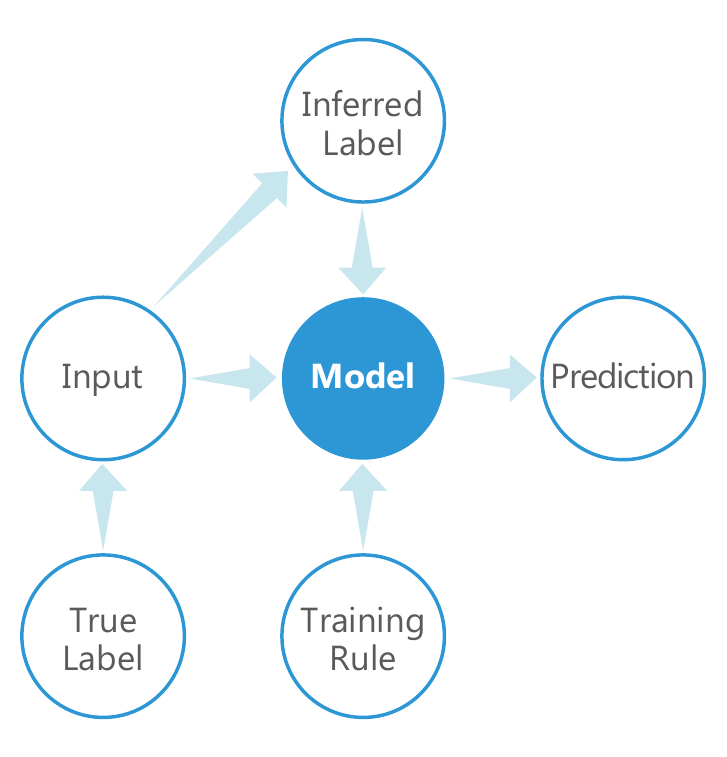} % first figure itself
        \caption{Causal model representation of the training. The input, the label and the training rules directly affect the model, whilst the model affects the final prediction. The Inferred Label and the True Label are also separate, while the former is given to the model and inferred from the input, the latter is the cause of the input and is not always available.}
        \label{fig:causalrelatiuonshops_1}
    \end{minipage}
    \hfill
    \begin{minipage}{0.45\textwidth}
        \centering
        \includegraphics[width=0.9\textwidth]{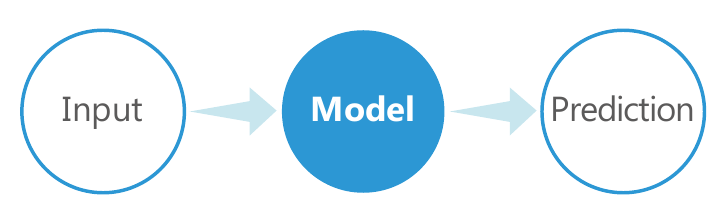} % second figure itself
        \caption{Causal model representation of the inference process. The Model node is a causal system by itself and it is the system that we are trying to understand.  }
        \label{fig:causalrelatiuonshops_2}
    \end{minipage}
    \label{fig:causalrelatiuonshops}
\end{figure}

The climb from the first to the second rung presents itself as a task with a diverse set of pathways. Differently from the direct explanation presented in previous chapters regarding interventions, this operation can take many forms when applied to artificial intelligence. In a sense, we can describe any of the operations that will be discussed as an intervention to the deep neural network that we want to describe in causal terms. Nonetheless, there are fundamental differences between them. Moreover, some of these interventions require careful analysis, as confounding bias is a threat that remains present~\cite{skelly2012assessing}. In contrast with the previous rung, the methods in this rung require their user to be, at least, partially aware of the causal relationships of the problem. Moreover, we further differentiate between the different elements in an AI system. Our overview of the system's causal structure can be seen in Figure~\ref{fig:causalrelatiuonshops_1}. It is important to distinguish between the inferred label and the true label. The former is defined by an human annotator and is inferred from the input itself. Moreover, it has a direct impact on the model. The latter is the true label, it is not always available and is the cause of the input, as discussed in medicine, the disease causes the image alternation and not the opposite~\cite{castro2020causality}. Hence, the latter has an indirect effect on the former too.

In the second rung we can have both post-hoc and in-model methods. Within each of these two groups it is also possible to find distinct strategies to do interventions, or to partially find a causal representation of a deep learning model. For the post-hoc interventions, we highlight interventions in the input and interventions in the prediction.

\begin{itemize}
    \item \textit{Intervention in the Input:} As shown in Figure~\ref{fig:causalrelatiuonshops_1}, the input given to the model has a direct arrow pointing towards the model. In other words, there is a causal effect between the input and the model, which means that changing the input will have consequences on the model's behaviour. It is possible that, by intervening on the Inputs by modifying them to the desired values, we can observe and learn the causal relationships between the model and the prediction. This type of interventions allows for the interpretability assessment to occur after training. It can be particularly useful if one aims to detect the implications of skin tone on the final prediction. A transformation in the skin tone of the image might reveal some hidden causal relationships between the model and the prediction, that should not exist. The relevance of this type of interventions is growing with the current developments of generative artificial intelligence methods, such as Diffusion Models~\cite{struct2023childnet,zhao2023diffswap}. 
    \item \textit{Intervention in the Prediction:} Explaining the model is the main goal of xAI, nonetheless, in some scenarios explaining the decision might represent a good proxy. As such, one of the most common strategies to explain these predictions relies on finding similar samples in the input space that produced similar predictions. This behaviour might help the human to focus on the common areas of both inputs. The most relevant implementations of these interventions are usually related to medical images~\cite{montenegro2023anonymizing}. An extension of these case-based explanations can be used as a proxy to find counterfactual samples. Setting the prediction as $P$, it is possible to find the nearest input, $X$, with a different value for $P$. The difference between both inputs highlights some reasons that distinguish both classes. 
\end{itemize}

On the other hand, for the in-model methods, we highlight the introduction of causal information in the model. This can be achieved through training or strong model constraints. The main advantage of these approaches is that they provide partial control of the model, which translates into an \textit{expected behaviour} that can be explained through the causal "rules" directly given by researchers.  In other words, as seen in Figure~\ref{fig:causalrelatiuonshops_2} acting on the Training Rule affects the model in a manner that can be predicted, and acting on the model also leads to predictable behaviour. This means that both approaches reveal a part of the causal model that will represent the final version of our system.

\begin{itemize}
    \item \textit{Model Priors:} Deep neural network architectures are, in the majority of the cases, shared across applications and tasks. This is possible due to their high expressiveness and capability to represent a myriad of complex functions. Nonetheless, this representativeness reinforces the opaqueness of the system. However, it is possible to insert certain decision choices in the model that reduce the function space, and lead to a predictable behaviour under certain circumstances. An example is the rotation equivariance on certain filters of the network~\cite{2023MargaridaGouveiaBIOSIGNALS,castro2023symmetry}. Another approach aligned with this strategy is the ProtoPNet~\cite{donnelly2022deformable,chen2019looks} which learns prototypes that can be used afterwards to explain model predictions. 
    \item \textit{Training Rule Priors:} In a similar fashion, these priors reveal certain parts of the causal system that represents the learnt deep neural network. However, these constraints are usually more relaxed than when applied to the model itself. Some examples are orthogonality constraints through the minimisation of the inner product of two vectors~\cite{Neto2022OrthMAD}, occlusion robust losses that promote similar embeddings between occluded and non-occluded samples~\cite{neto2021focusface}. Due to their relaxed nature, these priors are seen as an approach to promote a desired behaviour, usually a behaviour data researchers already know to be true. For that reason, and since they do not completely enforce the constraint, the interpretability of the model is usually lower than with Model Priors.
\end{itemize}

The range of approaches that fit the second rung of the causal explainability ladder is wide and might benefit from one or several of the previously described strategies. For instance, it is possible to have strong model priors, and leverage them through interventions on the predictions to construct a naive counterfactual. A naive counterfactual can be achieved if a system is capable of providing to a user a close and plausible alternative to the original input that belongs to a different class. It is frequently achieved through a search on the latent space~\cite{khorram2022cycle}. The term naive highlights the ignorance related to the inner relations of the model, and a focus on searching on the latent space. Some methods promote a certain structure on the latent space, and thus, this search can be somewhat controlled for certain variables. This introduces the capability to have more human friendly explanations. However, this is not easy to devise in practice due to humans' limited range of perception and control over the majority of the variables~\cite{Martins2016}. Thus, what a human searches for in a counterfactual explanation might not be fully represented in the provided explanation, leading to a trial and error search.

\subsubsection{The Third Rung: Imagining/Extrapolating}

Constructing and understanding the causal model that represents a deep neural network is not trivial. At the time of writing this article, there was no approach capable of fully explaining a deep neural network. As such, the third rung is yet to be achieved. The majority of the efforts have been devoted to the first two rungs, as such explanations suffice for the vast majority of tasks. Nonetheless, in critical domains, such as biometrics and healthcare applications, a incomplete knowledge of a deep learning model might prove itself to be a challenge in terms of security and user acceptance. 

The third rung of the explainability ladder is the last rung in the latter, which means that explanations provided by these methods are the highest possible degree of explanation. One very relevant example to compare this rung with the second rung is the construction of counterfactual samples. Differently from a naive counterfactual, a true counterfactual (or just counterfactual) originates from the complete understanding of the models' inner working mechanisms. Instead of searching on the latent space, one can wonder "What would be my prediction had my input image been given with a different head pose?". Moreover, this highlights the capabilities for detection and mitigation of biases, through simple queries, such as "Had the ethnicity been different, would it represent a worse prediction?". And we do not need to directly compute these values, as we can directly try and find the relationship between skin tone and certain predictions. 

Locating problems within a deep learning model is a particularly difficult task. However, when the causal model of that neural network is known, it is possible to quickly debug it and look for unwanted/spurious relationships through the comparison of that causal model and a partial model of the "world". And while some relationships might be spurious in real life, they can represent causal knowledge to the learnt deep neural network. Hence, highlighting information wrongly acquired by that same network.

%--------------------------------------------------------------------------------
\subsection{Work in xAI for Biometrics }%for computer vision that can benefit Biometrics and Categorization of works done: V / I / E}

%%\textcolor{red}{Dividing this subSection in the following two subsubsections looks better to me. I will wait for your feedback tho. }
% Attention Mechanisms
% Post-model explanations
% Multimodal?
%\textcolor{blue}{Topic: (examples of interesting/current/recent works and possible relationship with biometrics)}
%\textcolor{orange}{Table with some examples? }

Recently, the number of applications of methods that merge techniques of explainable artificial intelligence and biometrics has grown. As such, biometrics is a great case study for analyzing these techniques and how they fit within the proposed taxonomy. In this review, we analyzed 81 different papers published between 2018 and 2024. These studies have focused on distinct modalities and applications of biometrics with varying degrees of complexity. The curation of the paper list follows two main steps. First, a comprehensive keyword-guided search was conducted on papers indexed on Google Scholar. The keywords used are related to biometrics and explainable artificial intelligence. Subsequently, we manually scrapped the papers published at the main Computer Vision and Biometrics conferences, which were selected from the Google Scholar conference ranking\footnote{\url{https://scholar.google.com/citations?view_op=top_venues&hl=en&vq=eng_computervisionpatternrecognition}}. From the selected research documents, 53.1\% were published between 2022 and 2023, highlighting progress and interest in the field. In comparison, only 37.0\% of these studies were published between 2020 and 2021. A small proportion (approximately 9 \%)  was published between 2018 and 2019. As shown in Table~\ref{tab:xai4bio_works}, the face is the most frequent modality (76.5\% of the papers), with applications in recognition, morphing, presentation, and deepfake attack detection.  Owing to its easier acquisition and familiarity, the face is an obvious target for explainability research.  Iris recognition and presentation attack detection have also had several studies focusing on bringing explainability to these methods.

Nearly half of the studies analyzed did not include any mechanism to promote interpretability in their designs. They rely exclusively on explanation-based methods that are used in the vast majority of situations to produce visualizations and observational information of a potential explanation for a prediction.  Thus, 37 of the 81 papers were placed on the first rung of the xAI ladder. For instance, Trokielewicz~\textit{et al.} published studies on recognition~\cite{trokielewicz2019perception} and presentation attack detection~\cite{trokielewicz2018presentation} on cadaver irises. Both approaches rely exclusively on Grad-CAM to produce visualizations, similar to Hu~\textit{et al.}~\cite{hu2020end} for regular iris recognition. Nonetheless, not all visualizations are equal, and the approach used is an important element. Fu~\textit{et al.}~\cite{fu2022towards} introduced differential activation maps that were tweaked to explain demographic biases from the perspective of face recognition systems. The activation maps were grouped according to the ethnicity or gender of the individual, and the mean and variation information about them was extracted. These were used to construct a final map that illustrates the deviation from the Caucasian class.

Although extensively used for two-dimensional inputs, visualization methods are not restricted and can be applied to different inputs. For instance, Pinto and Cardoso~\cite{pinto2020xecg} studied different components of an ECG signal to determine the most relevant component for a recognition task. Four distinct visualization methods were used to dissect ECG signals. Similarly, Lim~\textit{et al.}~\cite{lim2022detecting} proposed the use of visualization techniques, such as Deep Taylor, to study the differences in rhythm and pitch between DeepFake and human voices, and LRP to uncover characteristics not seen through human visual perception. Deepfake applications rely mostly on approaches that belong to the first rung. Mazaheri and Roy-Chowdhury~\cite{mazaheri2022detection} proposed the use of CAM-based methods to detect inconsistencies in a manipulated image. Their method achieved state-of-the-art results for both DeepFake and face-manipulation detections. In the attack detection domain, Seibold~\textit{et al.}~\cite{seibold2021focused} and Xu~\textit{et al.}~\cite{xu2022supervised} presented methods for detecting morphing and deep fake attacks, respectively. The former proposed Focused Layer-wise Relevance Propagation (FLRP), which is an extension of Layer-wise Relevance Propagation (LRP), to explain the predictions of the network to a human. This new method focuses on intermediate layers to generate a visualization map so that it can capture the artefacts that characterize morphing attacks. Its use has proven to be particularly beneficial in situations where the backbone neural network displays uncertainty.  The novel contrastive solution proposed by Xu~\textit{et al.}~\cite{xu2022supervised} for deep fake detection can be generalized for distinct generators of attacks. More importantly, through the use of heat maps and Uniform Manifold Approximation and Projections (UMAP), it is possible to provide a simple visualization of the model decision. Genovese~\textit{et al.}~\cite{Genovese2019} introduced a method to simulate facial ageing. While the proposed generative method does not include any interpretability element, the authors proposed a novel method to explain the behaviour of generative adversarial (GAN) models. The method known as the Cross-GAN Filter Similarity Index (CGFSI) uses extracted filters from two generators, classifies them as belonging to the same or different categories and finally gives a similarity score for both generators. Sharma and Ross~\cite{sharma2021viability} tackled the problem of iris presentation attack detection (PAD). They proposed the integration of an Optical Coherence Tomograph (OCT) within the systems that aim to detect presentation attacks. Besides the promising results, the performance of the model was studied from the explainable artificial intelligence perspective. The t-distributed stochastic neighbour embedding (t-SNE) and the Grad-CAM algorithm are used to create a visualization of the prediction in a smaller latent space and to generate a heatmap to spatially highlight the most relevant areas for the prediction.

In contrast to CAM-based methods, LIME-based approaches intervene in the input through slight variations. These interventions allow for an effective assessment of the importance of certain features and how local changes in these features affect the final performance of the system. Although not as effective in images as it is in explaining interpretable features, Rajpal~\textit{et al.}~\cite{rajpal2023xai} proposed an LIME-based explanation for face recognition systems. This approach was tested on four distinct neural network architectures and three different datasets. However, the proposed methodology is not comparable to the current literature because of the chosen architectures and experimental design. Both Bousnina~\textit{et al.}~\cite{bousnina2023rise} and Lu and Ebrahimi~\cite{lu2023explanation} explored RISE-based~\cite{petsiuk2018rise} to explain face verification algorithms. The former generates randomly masked samples from the probe image and compares all samples to the reference image. Based on these comparisons, similar or dissimilar regions were highlighted for positive and negative matches, respectively. In the latter work, Lu and Ebrahimi~\cite{lu2023explanation} proposed a variation of Rise called S-RISE. Their approach consisted of comparing an unmasked probe with a masked reference image and a masked probe with an unmasked reference image. Based on these comparisons, it was possible to generate explanations that were evaluated quantitatively through insertion and deletion processes. Huber~\textit{et al.}~\cite{huber2024efficient} expanded on patch-replacement approaches as a form of intervention on the input. Their work focused on distinguishing between the contributions to the decision of the different features in the feature vector, and based on those individual contributions, it was possible to generate explanations that aligned with the final prediction. For positive matches, only similar features between the two images matter to find an explanation, whereas in negative cases, only dissimilar features matter. They further use this information and propose a new dataset called Patch-LFW, which allows them to assess explanations through the questions "What would be my explanation if I swapped this part of the face?". A similar patch-replacement strategy was studied by Knoche~\textit{et al.}~\cite{knoche2023explainable}, who also proposed an intervention on the prediction approach that was further expanded by Neto~\textit{et al.}~\cite{neto2023pic}. The latter approach generates an interpretable and probabilistic score for the final decision of a verification system based on genuine and impostor distributions.  

Interventions on the input have also been shown in fingerphoto verification and post-mortem iris recognition. Ramachandra and Li~\cite{ramachandra2023finger} tackled the complex task of fingerprint recognition through a mobile camera photo, and additionally, to the state-of-the-art performance, this study included several visualization techniques, such as Grad-CAM and gradient attribution, as well as interventions on the input such as LIME and Occlusion Sensitivity~\cite{zeiler2014visualizing}. 
 Boyd~\textit{et al.}~\cite{boyd2023human} expanded the explainability approaches to post-mortem iris recognition proposed by Kuehlkamp~\textit{et al.}~\cite{kuehlkamp2022interpretable} that is aware of the areas that are altered by the lack of blood flow. . Their new proposal included human annotators that indicated which areas of the two distinct iris images matched. Based on these annotations and features, the final algorithm was able to determine which areas were a strong match between two images, in accordance with human beliefs. Leveraging this information allows this method to be perfectly aligned with the expectations of the final users, which are forensic examiners.

While the majority of the previous approaches do not necessarily reflect a change in the training process of the deep neural network, some other methods have focused on introducing regularization, priors, and assumption constraints to their learning rule. Specifically, in Face Recognition,  Neto~\textit{et al.}~\cite{neto2021my,neto2021focusface} tackled the problem of masked face verification through the incorporation of additional terms in the loss that represent some domain knowledge previously known by the researcher. In this case, acknowledging that the mask is not relevant to the final prediction, the losses constrain the model output feature vector to be invariant to a potential mask. Huber~\cite{huber2021mask} presented a similar approach for designing a knowledge-distillation framework. Similarly, Franco~\textit{et al.} presented two studies ~\cite{franco2021learn,franco2022deep} that targeted the mitigation of biases within the face recognition system. Following the Demographic Parity constraint~\cite{calders2009building} the authors designed several regularization terms based on the Tikhonov/Rigde regularizer that were initially introduced by ONeto~\textit{et al.}~\cite{oneto2020exploiting}. Considering that a face has two distinct components, Han~\textit{et al.}~\cite{han2022personalized} developed a training rule in which a face generates a kernel that is subsequently disentangled into two distinct kernels. The first commonality component represents the shared characteristics among the subjects in the reference set used for optimization. The second component represents the special features of a certain person and is composed of the residuals between the difference of the initial kernel and the commonality kernel. This method showed a significant improvement over conventional CNNs on the three different datasets. Yin~\textit{et al.}~\cite{yin2019towards} presented one of the first approaches for interpretable face recognition. The proposed algorithm was optimized using two novel losses: Feature Activation Diversity (FAD) loss and Spatial Activation Diversity (SAD) loss. The former promotes filter robustness against occlusions, whereas the latter encourages the inclusion of semantic information. Finally, predictions can be analyzed by studying the locations of the filter responses.

Williford~\textit{et al.}~\cite{williford2020explainable} presented one of the first face recognition approaches that included interventions on both the input and predictions as well as a trained rule prior. They investigated a new form of explanation, based on triplets and an inpainting game. The main goal was to identify regions that were similar between the probe and the positive and dissimilar regions between the probe and the negative. Furthermore, during training, two other approaches to promote interpretability were studied: subtree Excitation Backpropagation (EBP) and Density-based Input Sampling for Explanation (DISE). The former computes the gradients of a triplet for every node of the original network and ranks them into a subtree. These are subsequently used to construct a saliency map that supports these explanations. The latter leverages these outputs to perform prior-based sampling of perturbations in the input image. This is an extension of a previous method known as Randomized Input Sampling for Explanation (RISE)~\cite{petsiuk2018rise}, which did not leverage any prior for the sampling. Later,  Liu~\textit{et al.}~\cite{liu2021heterogeneous} addressed the heterogeneous face recognition problem using adversarial training to disentangle different face modalities. Through this approach, their method, called heterogeneous face interpretable disentangled representation (HFIDR), can introduce semantic and interpretable information in the latent space. In this representation, information about identity and information related to modality are explicitly separated. Thus, conditioned on prediction, the author managed to reconstruct the original image in a different modality space. Recently, Terhörst~\textit{et al.}~\cite{terhorst2021pixel} used pixel quality indicators to explain face recognition systems. One of the most novel proposals of this approach is the adaptability of the method to different face recognition systems via the propagation of an input image 100 times through stochastic variation of the original network. The authors also proposed inpainting strategies to mitigate the impact of low-quality pixels in a face image. Although not completely interpretable pixel-quality can refer to obstacles in the recognition process, such as occlusions. 

Early studies on explainable AI applied to presentation attack detection (PAD) aimed to understand what makes an attack different from a genuine image. While it might be easier to check, given an appropriate clue, if an image is indeed an attack, it is not trivial to define the differences between attacks and genuine images. Jourabloo~\textit{et al.}~\cite{jourabloo2018face} attempted to model spoof information induced by each attack. For this, it was necessary to make certain assumptions regarding the properties of noise, such as its ubiquity. Through this method, it was possible to decompose the original image into noise and facial features. Liu~\textit{et al.}~\cite{liu2020disentangling} approached the facial PAD problem from a similar perspective. Their disentanglement approach was flexible enough to not only detect spoof and live samples through the disentangling of spoof traces but also to reconstruct a live version of that same input or synthesize a new spoof image with the same style and a different reference live image. Wang~\textit{et al.}~\cite{wang2022disentangled} method is another example of this type of approach. 
By decomposing the input image into a multi-scale frequency image with 12 channels, Fang~\textit{et al.}~\cite{fang2022learnable} achieved state-of-the-art results in cross-dataset scenarios. This ability to generalize aligns with the fact that the frequency representation is invariant to the dataset and that attacks display distinct frequency patterns. The architecture of the detection network was further designed using a hierarchical attention mechanism to support the detection of these different patterns~\cite{gonccalves2022survey}. Yang~\textit{et al.}~\cite{yang2019face} proposed a method that works with both temporal and spatial features. Specifically on the spatial features, the author splits a single image into several patches of the same image, which are processed by a shared-weight neural network. Subsequently, an attention module weighs the importance of each patch and provides a final classification. This training strategy favours an interpretable notion of the patches that influence the decision. Similarly, Bian~\textit{et al.}~\cite{bian2022learning} introduced features that can be interpreted by design. The proposed method was designed to work with both facial and environmental cues. However, the final decision uses a concatenation of interpretable and obscure features, making it difficult to trace the motives for the decision. Pan~\textit{et al.}~\cite{pan2022attention} proposed a framework with an integrated attention mechanism and generator for Grad-CAM visualization. The authors proposed the use of visualizations to optimize the training of the attention mechanism to which they designed a set of experimental studies and comparisons. Additionally, their method can generate textual explanations from a set of predefined text options.

To solve Morphing Attack Detection (MAD), and knowing that morphing samples hold two identities, whereas genuine samples hold a single identity, Neto~\textit{et al.}~\cite{Neto2022OrthMAD} designed a deep neural network that outputs two different feature vectors. These vectors have the particularity that their optimization includes a term to minimize the inner product. Thus, orthogonality is promoted and there is no mutual information between them. Simultaneously, they were concatenated and used to make the final prediction using an MLP layer. After identifying some flaws in previous work, Caldeira~\textit{et al.}~\cite{caldeira2023unveiling} proposed an improved version of the previous method. The orthogonality constraint was relaxed; instead, an autoencoder trained to reconstruct faces was used to create feature vectors. Hence, in the morphing scenarios, the training included two feature vectors of the original images produced by the autoencoder and two feature vectors produced by the proposed network from the morphed image. The new constraint consists of approximating the angle between the projections of each model. Additionally, owing to the intractability of the identity source from a concatenated vector, an interpretable score calculation was proposed. Each vector is used to predict the existence of an identity. When both indicated the presence of a face, the model flagged the input as a morphing sample.

Chen and Ross~\cite{chen2021} presented an attention-based approach to tackle iris presentation attack detection. When compared to previous attention-based their approach achieved significant improvements, which are reinforced by Grad-CAM generated visualizations. On the other side, Fang~\textit{et al.}~\cite{fang2021iris} developed a patch-based approach assisted by attention mechanisms to solve a similar task. The proposed approach has a loss that works at the pixel level as an auxiliary to the main task of detecting presentation attack inputs.  As such, it is also possible to obtain a pixel score or, if intended, a patch score considering a group of pixels. It works especially well when combined with spatial attention blocks, as demonstrated in the Score-CAM visualizations. Joshi~\textit{et al.}~\cite{Joshi2021} leveraged the stochastic nature of Monte Carlo dropout to generate uncertainty maps on the segmentation of the fingerprint regions of interest.

In fact, despite the visible growth of xAI applied for biometrics, some important elements are often discarded or ignored. No work explored the effects of causality on the production of explanations, nor the fact that rarely there is only one interpretation for a given explanation/visualization. Moreover, there is no focus on abnormal or rare events that can be extremely useful to produce explanations.  The usage of semantic explanations and the analysis of the human-friendliness of the explanations produced are nearly nonexistent. Thus, it is important to look into what is being done in other areas of computer vision regarding xAI, and how that work can be transposed or adapted into biometric applications. Furthermore, the implementation of these methods must be applied across the most diverse biometric traits and tasks. There is still a long path ahead for xAI methods applied to biometrics, and to introduce some consensus and uniformity between the needs of these methods. Despite a categorization of methods in the different rungs, not all the methods in a given rung are equally explainable, and it is important to pay attention to the methods that focus on going beyond simpler strategies, such as LIME.

\input{table_methods2}

\subsection{Work in xAI computer vision that can benefit biometrics}
% Interpretability / Explainability
% This Looks Like That: Deep Learning for Interpretable Image Recognition

The scenario of interpretable methods for biometrics has grown and highlights an increasing research interest in this domain. However, other computer vision domains had the opportunity to further develop and deeply explore different approaches to xAI. These approaches, independently of the rung where we place them, are, in most cases, yet to be applied to biometrics. Given its impact on medical image and other critical domains, it is undeniable that they might present themselves as attractive research directions for anyone looking towards improving the explainability of biometric systems. Hence, the following paragraphs delve into some roads already crossed in the computer vision literature. As seen in Table~\ref{tab:xai4bio_works}, one of the most common training rule constraints on second-rung approaches is the usage of attention mechanisms. These mechanisms have the advantage of being more interpretable, as they provide the importance of each element for their prediction. One of the most commonly used attention-based systems is the Transformer~\cite{vaswani2017attention}. This architecture led to advances in both computer vision and natural language processing domains. However, it is usually data-hungry, expensive to run on inference and not yet ready for embedded devices. Additionally, the recent usage of transformer-based foundation models further highlights the course of complexity and the opaqueness of these systems as they scale.

In 2019, Chen~\textit{et al.} proposed a novel deep neural network, \textit{prototypical part network} (ProtoPNet), that contains an inner reasoning process that may be considered interpretable due to its transparency~\cite{NEURIPS2019_adf7ee2d}. This model is trained in an end-to-end fashion and it is optimized to learn \textit{prototypes} (\textit{i.e.}, in this case, prototypes are latent representations) that are specific to a given class. To achieve this, the authors proposed to use generalized convolution by including a~\textit{prototype layer} that computes the squared $L_{2}$ distance instead of the inner product. Moreover, they also proposed to constrain each convolutional filter to be identical to some latent training patch. And thus, allowing us to interpret the convolutional filters as visualizable prototypical image parts. When the model is shown a new image, it dissects the image to find parts that may look like a prototypical part of some class and makes its prediction based on a weighted combination of the similarity scores between parts of the image and the learned prototypes. This approach is fitting to identification and authentication problems in the biometrics domain. Due to the capability to create patch-related explanations, it becomes easier to provide an understandable explanation to the user. Besides, this pipeline could help algorithm developers to assure that their models respect the ``right to explanation''~\cite{kaminski2019right} in the sense that this reasoning process is transparent to the end-user. Additionally it could be possible to find common and identity-related face elements. On the strongest challenges of these methods is their classification-based design, which makes them inappropriate for open-set tasks, such as face verification.  Other works have also explored the use of prototypical networks~\cite{hase2019interpretable,singh2021these,zhu2021low,nauta2021neural,ming2019interpretable}.

Inspired by the Recognition-By-Components cognitive psychology theory proposed by Biederman, Saralajew~\textit{et al.}~\cite{saralajew2019classification} proposed an architecture called Classification-By-Components network (CBC). The proposed network is optimized to learn generic characterization components for object detection. Moreover, it follows a class-wise reasoning approach with three distinct types of reasoning (positive, negative and indefinite) that is learnt to solve the classification task. The combination of the reasoning types branch into a probabilistic classifier and the decomposition of objects into generic components provides a precise interpretation of the decision process. This approach is even shown to be able of explaining the success behind an adversarial attack, which can be particularly useful for critical biometric systems. It could be explored if other forms of attacks to biometric systems can benefit from this approach. Other relevant application could be related to an extension of patch-based iris recognition work~\cite{boyd2023human}.

Qi~\textit{et al.}~\cite{qi2021embedding} proposed a novel Explanation Neural Network (XNN), which maps the output embedding into an explanation space. Their XNN algorithm is called Sparse Reconstruction Autoencoder (SRAE), and the produced explanation embedding is capable of retaining the prediction power of the original feature embedding. This explanation space can be understood and visualized by humans when employed a visualization approach proposed by the authors. Recently, Utkin~\textit{et al.}~\cite{utkin2021combining} proposed a similar approach based on autoencoders. Current works in biometrics, such as the one presented by Caldeira~\textit{et al.}~\cite{caldeira2023unveiling} could leverage this approach to further boost the explainability of the autoencoder in their proposed approach.

In 2021, Booth~\textit{et al.}~\cite{booth2021bayes} presented a novel model inspection method called BAYES-TREX. Instead of relying on potential abnormal behaviour in the test set, this method finds in-distribution examples with specified prediction confidence. This probabilistic method can reveal misclassifications that have high confidence and display the boundaries between the classes through ambiguous samples. Moreover, it can also predict the behaviour of the network when exposed to a sample from a novel class. The importance of the prediction score in verification tasks is of higher relevance. Some methods present scores that are difficult to interpret, and this type of approaches aligns with the proposals of Neto~\textit{et al.}~\cite{neto2023pic} and Knoche~\textit{et al.}~\cite{knoche2023explainable} to create an interpretable matching score.

Aiming at including some semantic meaning into the inspectability methods of deep neural networks, Losch~\textit{et al.}~\cite{losch2021semantic} proposed Semantic Bottlenecks (SB). These are integrated into pretrained models and align their channel outputs with visual concepts. The proposed method is rather impressive since even with a reduction of the number of channels by two orders of magnitude, the results are still on par with the state-of-the-art for segmentation challenges. SB have two variations, the unsupervised SBs (USB) and the concept-supervised SBs (SSB). Despite the similar performance, the latter shows a higher degree of inspectability.

% Concept whitening for interpretable image recognition 
Another challenge related to the training of deep neural networks is the learning and proper visualization of concepts that are important for a given task. There are several methods that attempt to see inside the hidden layers of DL models, however, they are applied~\textit{a posteriori} (\textit{i.e.}, post-hoc analysis), which may lead to misleading conclusions regarding the properties of the latent space~\cite{chen2020concept}. Therefore, to avoid relying on post-hoc analysis of neural networks, Chen~\textit{et al.}~\cite{chen2020concept} proposed a mechanism called \textit{concept whitening} (CW). This mechanism applies a whitening operation to the latent space (\textit{i.e.}, decorrelation and normalization) thus forcing the axes of the latent space to be aligned with known concepts of interest. The authors showed that this approach may be a reliable alternative to the use of batch normalization without hurting predictive performance. This methodology may be of interest in the field of PAD since it could open a different path for the comprehension of the concepts related to \textit{bona fide} and attack signals. Moreover, it could be used to align with soft-biometrics attributes, which would help the system to provide a semantic and meaningful  explanation based on terms that the end user is familiar with. 

Due to being, by nature, more interpretable than more complex models, the interest in simple models is gaining traction. However, training these models frequently leads to worse performance. Hence, researchers have investigated the possibility to change how these models learn. One of the most used approaches, is the distillation of knowledge learned by a complex model (teacher) into a smaller one (student)~\cite{liu2018improving,alharbi2021learning}. It has been already been leveraged in biometrics, not as a promoter of interpretability, but as an auxiliary process for a different task~\cite{huber2021mask,boutros2022template}. Hence, even if the training of inherently interpretable models is less stable, it is possible to optimise it by gathering information from a more complex, opaque model. %% Citation Eduarda's survey in the future? 

Other methods that are being explored and can contribute to the increase of interpretability of biometric methods include counterfactual explanations and influence functions. And there is some work in that direction~\cite{jung2020counterfactual,wu2021polyjuice,hvilshoj2021ecinn,guo2020fastif}. However, they are still growing and maturing. In their current form, they are similar to the naive counterfactuals described in previous chapters as a second rung approach.  It is clear that xAI was already properly introduced into computer vision systems. Nonetheless, when comparing the variability of approaches presented in this Section with the ones shown in Table~\ref{tab:xai4bio_works} that biometric systems are far from being interpretable. Hence, this work also aims at motivating the need to accelerate the research of these methods so that biometric applications can reach the same level of interpretability shown by other computer vision fields.

There has been also attempts to build explainability pipelines capable of handling different types of machine learning models and distinct applications. IBM, presented ``The AI 360 Toolkit''\footnote{https://developer.ibm.com/articles/the-ai-360-toolkit-ai-models-explained/}, which is one of the most advanced pipelines for model and application agnostic explanations. Despite the potential of these systems, they lack the domain information and design that is frequently needed for complete explanations. Hence, while it can be seen as a good starting point, it must not be seen as the definitive solution.

Conducting a careful analysis of the methods presented in this section, it is possible to categorise them accordingly to the xAI ladder proposed in previous sections. As such, it shows the robustness of this new taxonomy across multiple domains. Nonetheless, it is also visible that the field is still far from being saturated and further efforts are needed to achieve the third rung where researchers can understand the causal relations in their models. 
%\vspace{4cm}

%%%%%%%%%%%%%%%%%%%%%%%%%%%%%%%%%%%%%%%%%%
\section{Illustrating the Benefits of Interpretable Biometrics}
\label{sec:four}
Interpretable biometrics are still being used exclusively for academic purposes. However, leveraging these techniques can impact lives outside the academic domain and, for instance, support vulnerable groups against discriminatory behaviour. Its integration with the systems that are currently deployed is crucial to the progression towards a more transparent artificial intelligence. Hence, this Section discusses the situations where interpretable biometric systems should be deployed, the benefits that shall arise from their usage, the limitations of the current approaches, the needs for innovation, and finally, it wraps up with some potential research directions.

%\textcolor{orange}{(Why are face recognition methods not good for face presentation attack detection? Shouldn't a method that recognizes and individual be capable of understanding what makes an human Human).
%\\Are we currently solving the right problems? Should we ask different research questions? (training process, evaluation, metrics)}. \textcolor{orange}{(Can we create a table with metrics and their "useful" information or something like weak and strong aspects of each metric)}

%\textcolor{red}{With the following 3 images we go through training processes, which questions to ask and how informative are current metrics. Needs to be double checked and extended of course.}

\subsection{Applications}

The applications of biometric systems are vast, and thus, the applications of their interpretable counterpart are also vast. This raises the question of how it is possible to create a single interpretable system for all the tasks. Realistically, this is a question that shall not be answered for now. Before being solved holistically, interpretable biometrics must be tackled at each smaller subdomain. Progression into a more general approach occurs iteratively through the explanation of the different methods. It can, for instance,  explore the validation of an interpretable approach on several tasks or traits. It is however essential to keep in mind that although it would be desirable to be an interpretable system that fits the most variate scenarios, in practice, it is known that there are no free lunches. Moreover, to achieve the third rung it is important to progress towards more complex methods within the second rung instead of focusing on the more simple approaches seen frequently. Besides the benefits previously mentioned, it is possible to discuss the impact of these systems in more areas. For this analysis, interpretable methods for biometrics are deconstructed into three areas where they can be integrated.

\subsubsection{Improving evaluation}

Metrics have been one of the most crucial elements for the raise of deep learning systems. Not only they are responsible for the understanding of how the systems are performing, but they are also used to establish comparisons between different methods. Moreover, measures such as the accuracy and model's confidence have been the golden standard to increase the public acceptance and deployment of these systems. Metrics are useful for and provide a significant portion of information, nonetheless, they also lack significant details. 

\begin{figure}[!h]
\centering
\includegraphics[width=\linewidth]{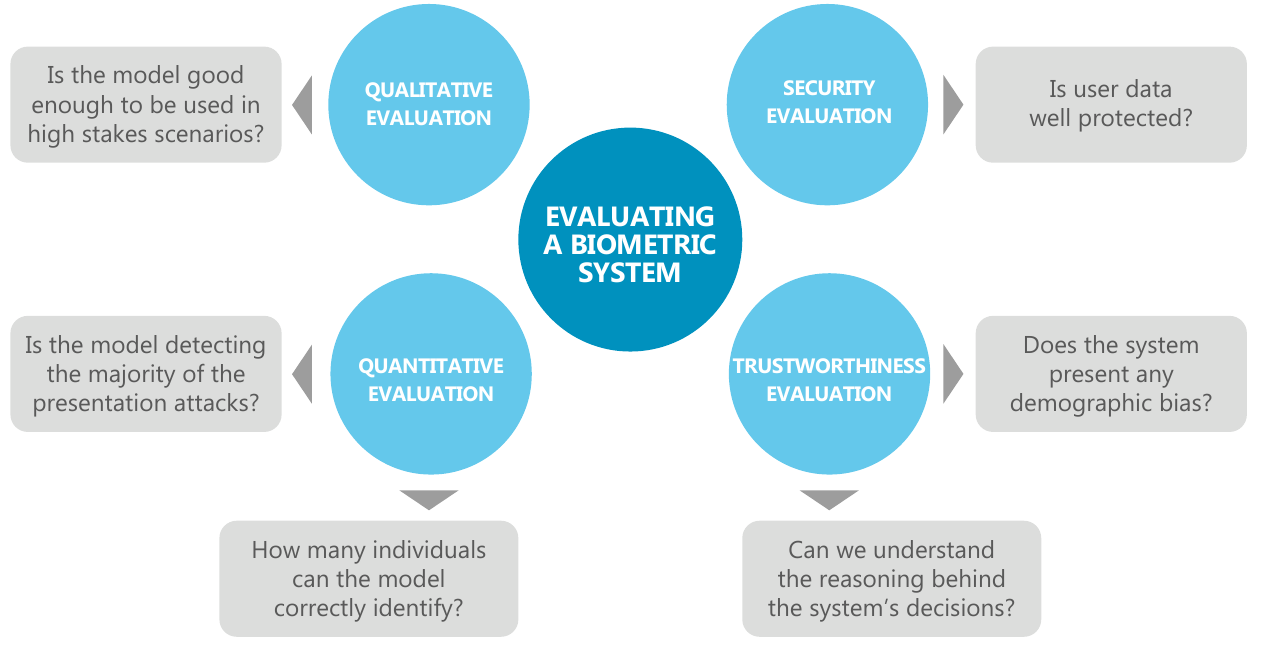}
\caption{Proposed structure for a complete evaluation of the performance of biometric systems at four distinct levels. Quantitative, qualitative, security and trustworthiness evaluation. The latter is focused on the insights provided by explainable artificial intelligence methods.}
\label{new_evaluation_analysis}
\end{figure}

Understanding how many predictions were correct and by which margin is needed for a quick analysis. However, a correct prediction might be correct for the wrong reasons, and a wrong prediction might be wrong for even worse motives. Hence, providing a single prediction with no explanation does not suffice to fully understand these hidden motives, it is necessary to show why. As such, the evaluation of deep neural network systems is far from complete, leading to flawed assessment of their performance. Leveraging our knowledge of the causal structure, attained through xAI methods, of the model that we have just built holds a higher potential for performance assessment. It allows for two types of evaluation. Retrospective evaluation where after getting a prediction, we try to find the "cause" of that prediction through our knowledge of the model. And also \textit{apriori} evaluation, where researchers can inspect the learnt rules and infer if they are aligned with the expected rules.

We propose an evaluation of deep learning systems that can be divided into four major focal points as displayed in Fig.~\ref{new_evaluation_analysis}. First, it is essential to quantitatively evaluate our system, and for this we use our typical metrics such as accuracy, F1-Score and others. It is of no use to explain a "dumb" model. Secondly, one should assess in a qualitative manner the performance of its model. In that sense it is important to understand why the model takes the decisions that it takes, how he reaches the decision and which factors are important. This leads to a third evaluation process where we, now fuelled with the knowledge of the "why's" of our system, inspect for bias and trustworthiness problems. It is worth noting that Huber~\textit{et al.}~\cite{huber2023explainability} has shown that certain Rung 1 approaches suffer from biases themselves. Finally, it is essential to understand the implications of explainability on the privacy of the users and the privacy of the system itself. For instance, Matulionyte~\cite{matulionyte2022reconciling} explored the implications that having a transparent system has on trade-secrets and how it can be possible to navigate in the trade-off between these two concepts. Additionally, counterfactual explanations have severe risks if not done properly. It could be possible that by proving such an explanation for a biometric recognition system we compromise the safety and privacy of another user.

\subsubsection{Providing feedback}

Whilst expressing different information, metrics and losses share a lot of similar details. The former, as previously described, supports the evaluation of deep learning models. While the latter provides information for the model optimization. Usually, the minimization of the latter improves the metrics. However, as already mentioned, a correct prediction might rely on incorrect cues. Hence, it is also important to have explainable artificial intelligence methods at the optimization level.

\begin{figure}[!h]
\centering
\includegraphics[width=5.5in]{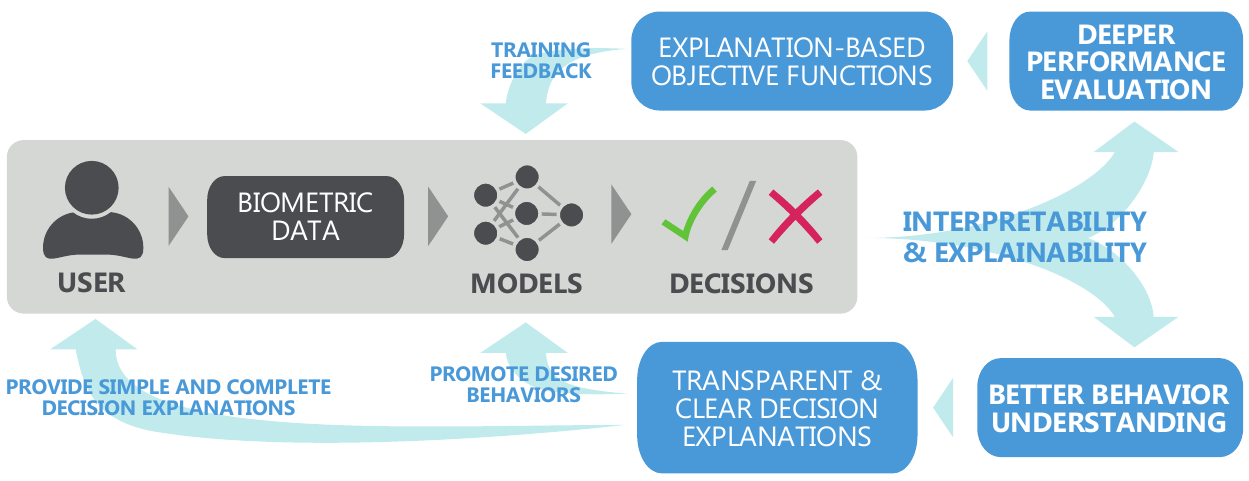}
\caption{Structure of a biometric system that leverages explainable artificial intelligence methods to improve its optimization function and to provide useful feedback to the user. }
\label{evaluation_scheme}
\end{figure}

The integration of xAI and optimization processes can be done in two ways, as shown in Fig.~\ref{evaluation_scheme}. First, it is possible to promote certain behaviours or properties through the implementation of constraints to the model. These constraints can be applied to specific layers or to the output of any layer. The second approach is to design objective functions that incorporate feedback from the explanations produced from the predicted batch. This latter approach is significantly more challenging. Nonetheless, there are works aiming to tackle this fusion. For instance, the framework proposed by Pan~\textit{et al.}~\cite{pan2022attention} tunes the weights of the attention mechanism through the visual explanations produced in the previous forward pass. Although it is a young field, these strategies have the benefit to directly affect the causal model learnt by the system. And, since we can control these strategies we can promote certain information to be learnt by our system leading to a lower knowledge gap between researcher and deep neural network.

\subsubsection{Protecting privacy}

Some approaches of explainable machine learning focus on image retrieval tasks. For instance, when explaining why two images do not match, it is possible to find examples of mated and non-matted images to create some sort of counterfactual explanation. However, this might expose users' private information. Silva~\textit{et al.}~\cite{silva2020interpretability} leveraged interpretability methods to guide the search for the appropriate image on a similar task on the medical image domain. Further research on the medical domain also focused on the privacy-preserving component of image retrieval~\cite{montenegro2021privacy}. A potential extension for biometric applications was already briefly addressed~\cite{montenegro2022privacy}. We argue that privacy-preserving methods are of extreme importance due to the nature of biometric data. Furthermore, it ensures that users' trust and public opinion, which are crucial for the applicability of biometric systems, remain untouched. Hence, the exposure and potential leakage of important and private information is minimised. The integration of these mechanisms in real-world situations depends on the desired degree of privacy and the efforts required to change the system.

%\textcolor{red}{Should we explain the metrics in the text or is it fine like that in the table? }

% \textcolor{blue}{Topics: Matching / Classification / PAD / Deepfakes det / Privacy protection }

% \subsubsection{Classification}

% \subsubsection{Template Matching}

% \subsubsection{Presentation Attack Detection}

% \subsubsection{Deep Fake Detection}

% \subsubsection{Privacy Protection}

\subsection{The incompleteness of the current explanations}

% Saliency Maps

\begin{figure}[!h]
\centering
\includegraphics[width=4.6in]{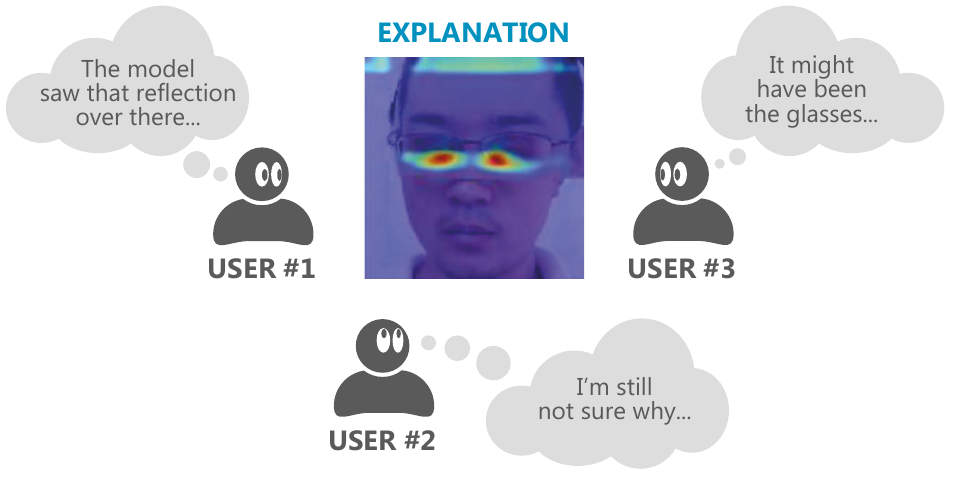}
\caption{Simple draft of a diagram exemplifying how Grad-CAM output as heatmap-based images can be interpreted in a different manner by humans.}
\label{fig:grad_cam_example}
\end{figure}

Rung 1 approaches such as saliency maps have been widely used to determine which parts of the image are being used by DL-models to perform a decision. However, it is important to acknowledge that knowing where the model is looking within the image does not tell the user what it is doing with that part of the image. A representative example of this is that we can obtain saliency maps for different classes, even if they have no relation to the true label. From all the "possible" explanations given to each class, only the predicted class maps are given to the user. This might create a false sensation of confidence in the decision, which might not be reflected in the overall explanations. For instance, the case where the explanations for multiple (or all) of the classes are identical. Moreover, these methods focus on presenting to the user the meaningful pixels, allowing the user to decide the reason behind the prediction based on those pixels. This raises a problem as humans are independent and subjective animals, and thus, from a simple heatmap \textit{N} different users can find \textit{N} different motives for the highlight of those pixels (Fig.~\ref{fig:grad_cam_example}). In a sense, Grad-CAM and similar methods give a small hint to the user regarding to where the model was looking at. The reason why those were the meaningful pixels remains unanswered and the prediction remains unexplained. Still from Figure~\ref{fig:grad_cam_example}, the model might have hidden biases against a demographic group, which are not carefully described by the observational explanation.

% Textual and Multimodal Explanations
Several models have been designed to generate textual explanations as part of their training process~\cite{gilpin2018explaining}. This approach has demonstrated interesting results in systems of visual question answering~\cite{antol2015vqa} and in algorithms for image classification~\cite{hendricks2016generating}. To achieve human-understandable explanations, the modules that generate explanations are trained on large data sets of human-written explanations, thus generating models that explain their decisions using natural language~\cite{hendricks2019visual}. This approach has been extended to multi-modal explanations, in~\cite{park2018multimodal}. In this use case, the model has an extra module (based on attention mechanisms) that generates visual explanations. Rio-Torto~\textit{et al.}~\cite{rio-torto2024parameterefficient} recently explored parameter efficient strategies to generate textual explanations for chest x-rays, leveraging the recently popular transformer architecture. There is, however, an important detail that we should take into account: the explanations are generated based on the decision, after the decision of the network has already been made. Hence, given that the explanations are learned in a supervised manner and that the generated explanations are conditioned by the output of the network, should we consider these as real explanations? What differs these models from general classification models? How can we be sure that the textual explanation that was generated truly represents the reasoning process used by the model to output a prediction? Additionally, the current concern regarding hallucinations of large language models highlights some potential problems of explaining an obscure system through another obscure system~\cite{xu2024hallucination, ji2023towards}.

\subsection{Unveiling biases}

Biometric systems have at their disposal inputs that frequently encode demographic information or sensitive attributes. Not only that, but it has been shown that these attributes are also encoded in the latent space of these models. And while this information is useful for increased performance, one must be careful with the way it is used. On very important topic, especially in face recognition, is the measurement and mitigation of racial bias. This bias is characterised by an intra-ethnicity profile, in other words it relates to the capability of correctly matching the individuals within an ethnic group. However, if these embeddings are used for different tasks, for instance, as a secondary input to a credit approval system, the hidden demographic information can lead to an increased biased inter-ethnicity hindering the necessity to explain the information contained in the latent space. 

While they can be measured, it is hard to directly infer the existence of biases simply by analysing the trained model. As such, several models are flagged as biased several days, weeks or even months after being released. This further reinforces the need for explainable AI, and through the proposed taxonomy we can see that the more causal relations of a model we understand, the easier it is to compare them to the causal relations of the \textit{real world}. Through this comparison we can easily find relations that are not aligned and that could represent a source of bias, such as the usage of the skin tone to predict crime re-incidence or credit assessment. Kei \textit{et al.} also argues that explanations can provide much-needed help in this matter~\cite{keil2004lies}. It is important to detect and act quickly on those cases. Frequently edge cases are related to under-represented minorities in the dataset~\cite{pmlr-v81-buolamwini18a}. Moreover, some datasets with specific data from some minority include lower-quality samples~\cite{vangara2019characterizing}. Hence, they are the most exposed to potential harms by algorithmic mistakes. Explanations can serve as a tool to mitigate these nefarious effects and impel machine learning into fairness~\cite{doshi2017towards}. Terhörst~\textit{et al.}~\cite{terhorst2021pixel} and Boutros~\textit{et al.}~\cite{boutros2023cr} are examples of these quality assessments that can be used to construct balanced datasets.

With the rise of deep learning-based biometric systems developed with direct applications to police forces and law enforcement, unveiling bias becomes mandatory~\cite{Almeida2016,Prema2019,Abomhara2020,Furberg2017}. Hence, these use cases have a high requirement for appropriate explanations and not incomplete ones. For instance, if a face presentation attack detection algorithm suffers from a demographic bias that is prominent on darker skin tones, it cannot be detected by visualization methods. Nor can it explain that the skin tone played a role in the decision. Moreover, if the explanation was given to a human, he would most likely focus his attention on the shape of the areas around the highlighted pixels, search for blurriness or artifacts. Hence, for the safe deployment of these systems one must attempt to learn the hidden rules that guide the model towards a decision. However, despite not being a complete explanation nor unveiling the hidden mechanisms of a deep neural network, rung 1 approaches are still useful for preliminary verification of the deep learning system.

%%%%%%%%%%%%%%%%%%%%%%%%%%%%%%%%%%%%%%%
\section{Conclusion: In which direction should we look in the future?}
\label{sec:five}

Throughout this document we have covered several topics in explainable AI. More importantly, we have covered a novel taxonomy for explainable artificial intelligence that benefits from the experience and insight of causality. The proposed taxonomy borrows concepts from the Ladder of Causality, leading to our proposed Ladder of xAI. In this, we have proposed a division of xAI method in three different rungs, where we consider that every deep learning model learnt has an inherent causal model that can represent its inner rules. The first rung, considers the simplest approach to xAI, visualisation methods, which bring observational information to the user. However, these methods are the equivalent of correlation for causality, as they answer a different question and are not guaranteed to be related to the model inner workings. On the second rung the number of different approaches grows significantly. First, we have approaches based on inserting prior to either the model or the training rule. Model priors, due to their strictness partially reveals the final causal representation of the learnt model, whereas training priors, due to its relaxed nature leads to a less effective revelation of that same model. Additionally, it is possible to make interventions at input and prediction levels. When there is an intervention at the input level, the researcher aims to know "How will this change affect my model predictions?". Future methods on the third rung will allow to answer this question immediately, and further answer to the "Why is this the behaviour of my model?". However, we are still far from reaching the third rung with deep learning-based methods.  

The xAI ladder was applied to 81 different research papers published between 2018 and 2024. These papers proposed a myriad approaches from the first two rungs for different biometric modalities and tasks. From Face Recognition to Post-Mortem Iris Recognition and DeepFake Detection, the range of tasks that benefited from explainable approaches is outstanding. Moreover, there seems to be a growth trend. As such, it is important that researchers have a common language to address, categorise and assess explainable methods.  Despite the current demonstration and application with biometrics works, the proposed taxonomy is sufficiently general to be applied to any domain of computer vision. Hence, other areas such as medical applications can benefit from this framework and converge to common ground. Additionally, this framework can be extended to fit the necessities of a specific domain as it is flexible and not bonded to a single strategy.

Additionally we have further discussed some points regarding explainable AI and how it could be used to improve different areas of biometrics. These discussed areas are frequently not seen as fields that can benefit from xAI, and for this we decided to illustrate some benefits that can be drawn from xAI approaches. In this sense, we have discussed the evaluation of models, the feedback given and the privacy protection of both user data and the trade-secrets behind an artificial intelligence model. Furthermore, we have discussed a few topics on unveiling biases with explainable AI. With the proposed taxonomy and strategy to approach xAI, the unveiling of biases begins during development and model assessment, reducing the amount of models deployed with hidden biases. 

\subsection{Future work}

Despite the obvious progress, there are areas, specially in biometrics that were not fully covered in this survey. As such, we provide a glance on some relevant future worrk topics that might provide novel strategies in xAI, safer biometric systems or trustworthy models.  And with the recent growing interest in artificial intelligence, mostly supported by the deployment of Large Language Models and Text-to-Image systems, it is of crucial importance that we remain aware of the dangers of these models, their misuse, and the strategies to mitigate them.

\subsubsection{Revisiting fundamental research in mathematics and computer science: Can we optimize our algorithms more efficiently?}

Although a clear interpretation and analysis of some of the older machine learning algorithms was somewhat possible, the majority of that clearness has been lost with the deep learning uprising. It is not trivial to reformulate these solutions in order to decrease its opaqueness. However, as seen in the previous Sections, there are already researchers working on variations of architectures and losses that promote a more controlled behavior. In an effort towards interpretable models, or models which can benefit from more complete explanations, it is crucial to redesign the current theory and algorithms with xAI in the scope. 

\subsubsection{Benefiting from multiple data sources: Can we build methodologies that take into account multi-modal data?
}

Working with multi-modal data might improve the performance of a unimodal method. However, this can be both a gift and a curse. When carefully implemented and designed, the integration of an additional data source can support explanations that are less ambiguous and more complete. Nonetheless, deep learning systems working on one modality are already increasingly complex, if an additional data source is incorporated, it might lead to more opaqueness. Hence, besides studying how to integrate multiple modalities, it is also extremely important to understand the compatibility of those modalities. The utilization of intrinsically interpretable models supports growth in the number of fused modalities without compromising the understanding of the model's inner workings.

\subsubsection{What to believe: On the coherence of the explanations with human perception}

Humans are creatures of beliefs and prior knowledge, and despite their potential misalignment with reality, going against them creates a lot of entropy. And explanations are, for now, limited to not being capable of counterarguing someone that disagrees with it. Hence, an explanation that presents a different view than the one already acquired by the user might suffer from strong mistrust.   Sometimes, this different perspective is on elements that are not the main focus of the explanation. And thus, to mitigate this, it is necessary to give explanations that also align with the users' view. To achieve this, it is necessary to craft the explanation method for the specific target user that is going to benefit from the machine learning model. Similarly, the subjectivity aspect of the explanations must be tolerated in certain use cases, whereas others require objective and precise explanations. Joining these concepts is not easy, and there is a significant amount of work to be done in order to direct the explanations to the target users. 

%\textcolor{blue}{Topic: Subjectivity of explanations / Coherence of explanations with human perception}

\subsubsection{Learning information: Correlation vs Causation}

Deep learning-based models are strong and complex function approximators also known as universal approximators. However, this complexity is not always beneficial. For instance, some background elements might happen more frequently on certain predictions, nonetheless, they are not the cause of that label. Learning these correlated relationships increases the chance that the network makes the inference based on them. The current framework, considers that each learnt model has an inherent causal model that explains its predictions, however, the learnt model is not guaranteed to be causal with the real world. In this sense, these correlations might represent causal knowledge when seen through the lenses of the deep model, potentially aggravating biases. An example of these is a face presentation attack detection system trained on images of people on a variety of environments and weathers. As shown by demographic information certain geographic areas are related to certain ethnic groups. A potential harmful correlation would be to classify images of people from certain ethnic groups as presentation attacks, just because the background is considerably different from the one used in the training data for the majority of that ethnic group. This relationship would present a causal rule within our learnt model that does not reflect the causal reality of the world. 

Focusing on causes of a certain effect leads to more faithful and believable predictions. Therefore, there is a research direction on causal learning and causal inference. It is of utter importance to integrate these practices in biometric systems and biometric research. Recently, there were reported several cases of biased biometric systems. These biases can be unveiled by xAI and mitigated with causal inference. Hence, this research direction must continue to grow in future iterations of biometric systems. 

\subsubsection{Intrinsically-interpretable models vs explanation methods}

While explanations methods represent the majority of xAI application in the biometrics domains, they do not represent a definitve solution for the black box problem. On the other hand, while intrinsically interpretable models, which are specifically designed to promote interpretability, are great tools to reduce the opaqueness of a model, they can still benefit of explanation generators. We believe that the future of explainable artificial intelligence relies in the joint work of in- and post-model approaches. It can be done through the development of differential post explanations that can provide feedback to the model, or through the design of explanation methods that rely on the intrinsic properties of a model.
%\textcolor{blue}{Topic: }

\begin{acks}

Nothing yet.
\end{acks}

\bibliographystyle{ACM-Reference-Format}
\bibliography{sample-base}

\appendix

\end{document}

%% file: table_methods2.tex
\begin{landscape}
\begin{center}\begin{small}
\begin{longtable}{|p{0.22\textheight}p{0.02\textheight}p{0.25\textheight}p{0.25\textheight}p{0.35\textheight}|}

    \caption{Methods that leverage explainability methods to inspect the predictions of biometric systems. Some of these methods, promote interpretability through the usage of certain techniques, whereas others focus mostly on explaining the predictions. Each method is placed in the respective rung in the xAI ladder. For the methods of the second rung they are further categorised into Model Prior (M. P.), Training Rule Prior (T. R. P.), Intervention on Input (I. I.) and Intervention on the Prediction (I. P.).}\label{tab:xai4bio_works}\\
    \toprule
    \textbf{Method} & \textbf{Year} & \textbf{Modality} & \textbf{xAI Ladder} & \textbf{Approach to xAI} \\
    \hline
    \endhead

    %\endfoot
    
    \multicolumn{5}{|c|}{\makecell{\textbf{Face Recognition}}} \\
    \hline
    Huber~\textit{et al.}~\cite{huber2024efficient} & 2024 & Face Verification & Rung 2 - I. I. + I. P. & Decision-based Patch Replacement and Individual Dimension Contribution to Decision\\
    Li~\cite{li2023bionet} & 2023 & Face Recognition & Rung 2 - M. P. & Biologically Inspired Architecture + Attribute Weight for Decision\\
    Correia~\textit{et al.}~\cite{correia2023face} & 2023 & Face Verification & Rung 2 - M. P. & Attention + Explainability Masks Construction\\
    Bousnina~\textit{et al.}~\cite{bousnina2023rise} & 2023 & Face Verification & Rung 2 - I. I. & RISE-Based Similarity/Dissimilarity Maps\\
    Knoche~\textit{et al.}~\cite{knoche2023explainable} & 2023 & Face Verification & Rung 2 - I. I. + I. P. & Patch Replacement\\
    Lu and Ebrahimi~\cite{lu2023explanation} & 2023 & Face Recognition & Rung 2 - I. I. & S-Rise\\
    Rajpal~\textit{et al.}~\cite{rajpal2023xai} & 2023 & Face Recognition  & Rung 2 - I. I. & LIME Maps \\ % is lime only interpretable when we have interpertable features?????
    Neto~\textit{et al.}~\cite{neto2023pic} & 2023 & Face Recognition & Rung 2 - I. P. & Interpretable Match Score \\
    Rodriguez~\textit{et al.}~\cite{rodriguez2023multi} & 2023 & Quality Face Recognition & Rung 2 - T. R. P. & Enviromental Attributes \\
    Terhörst~\textit{et al.}~\cite{terhorst2021pixel} & 2023 & Face quality for Face Recognition & Rung 2 - I. I. + I. P. & Pixel-level Quality Maps \\
    Fu~\textit{et al.}~\cite{fu2022towards} & 2022 & Face Recognition Bias & Rung 1 & Differential Activation Maps \\
    Han~\textit{et al.}~\cite{han2022personalized} & 2022 & Face Recognition & Rung 2 - T. R. P. & Personalized Kernels + Feature Maps \\

    Roy~\textit{et al.}~\cite{roy2022interpretable} & 2022 & Heterogeneous Face Recognition & Rung 2 - M. P. + I. I. & Invariant Features\\
    Winter~\textit{et al.}~\cite{winter2022demystifying} & 2022 & Face Recognition & Rung 2 - T. R. P. + I. P.  & Local Patch-based Features with Eexplainable Boosting Machine (EMB)\\
    Mery~\textit{et al.}~\cite{mery2022black} & 2022 & Face Verification & Rung 1 & Saliency maps AVG contours \\
     Fu~\textit{et al.}~\cite{fu2022explainability} & 2022 & Face quality for Face recogniton & Rung 1 & Score-CAM\cite{wang2020score} + Activation Maps\\
     Franco~\textit{et al.}~\cite{franco2022deep} & 2022 & Face Recognition & Rung 2 - T. R. P. & Tikhonov regularizers \\

    Neto~\textit{et al.}~\cite{neto2021my,neto2021focusface} & 2021 & Masked Face Recognition & Rung 2 - T. R. P.  + I.I. & Mask-Invariant Loss + Grad-CAM Visualisation\\
    Huber~\textit{et al.}~\cite{huber2021mask} & 2021 & Masked Face Recognition & Rung 2 - T. R. P.  + I.I. & Mask-Invariant Knowledge Distillation\\

     Franco~\textit{et al.}~\cite{franco2021learn} & 2021 & Face Recognition & Rung 2 - T. R. P. & Tikhonov regularizer\\
     Liu~\textit{et al.}~\cite{liu2021heterogeneous} & 2021 & Heterogeneous Face Recognition & Rung 2 - T. R. P. + I. P. & Modality/Identity Disentangled Representation + Cross-modality Synthesis \\
     Anghelone~\textit{et al.}~\cite{anghelone2021explainable} & 2021 & Face Recognition & Rung 1 & Heatmap-based Visualisation  \\
     Jiang and Zeng~\cite{jiang2021explainable} & 2021 & Face Recognition & Rung 2 - T. R. P.  & Face Components Representation and Self-Attention based Reconstruction + Components Similarity Matrix \\
        \hline

    Williford~\textit{et al.}~\cite{williford2020explainable} & 2020 & Face Recognition & Rung 2 - I.I. + T. R. P. + I. P. & DISE + subtree EBP +  Inpainting game \\
    Zee~\textit{et al.}~\cite{zee2019enhancing} & 2019 & Face Recognition & Rung 1 & Guided Backpropagation \& CAM \\
    Yin~\textit{et al.}~\cite{yin2019towards} & 2019 & Face Recognition &  Rung 2 - T. R. P. + I. I. & Spatial and Feature Activation Diversity Losses \\

    \hline
    
    \multicolumn{5}{|c|}{\makecell{\textbf{Face Morphing}}} \\
    \hline
    Caldeira~\textit{et al.}~\cite{caldeira2023unveiling} & 2023 & Face Morphing Detection & Rung 2 - M. P. + T. R. P. & Interpretable Score + Identity Disentanglement\\
    Dargaud~\textit{et al.}~\cite{dargaud2023principal} & 2023 & Face Morphing Detection & Rung 1 & PCA-based Visualisation \\
    Dwivedi~\textit{et al.}~\cite{dwivedi2023efficient} & 2023 & Face Morphing Detection & Rung 1 & Ensemble of Visualisations \\
    Neto~\textit{et al.}~\cite{Neto2022OrthMAD} & 2022 & Face Morphing Detection & Rung 2 - T. R. P. & Orthogonal Identity Disentanglement\\
    Myhrvold~\textit{et al.}~\cite{myhrvold2022explainable} & 2022 & Face Morphing Detection & Rung 1 & LRP \\
    Seibold~\textit{et al.}~\cite{seibold2021focused}& 2021 & Face Morphing Attack Detection & Rung 1 & Focused LRP \\

    \multicolumn{5}{|c|}{\makecell{\textbf{Face PAD}}} \\
    \hline
    Prasad~\textit{et al.}~\cite{prasad2022self} & 2023 & Face PAD & Rung 2 - T. R. P. & Intermediate Depth Estimation\\
    Pan~\textit{et al.}~\cite{pan2022attention} & 2022 & Face PAD & Rung 2 - T. R. P. &  Grad-CAM + Textual Explanations\\
    Fang~\textit{et al.}~\cite{fang2022learnable} & 2022 & Face PAD & Rung 2 - T. R. P. & Multi-Scale Frequency Decomposition + Hierarchical Attention + t-SNE \\
    
     Bian~\textit{et al.}~\cite{bian2022learning} & 2022 & Face PAD & Rung 2 - T. R. P. & Auxiliary Learning with Interpretable Cues \\% not 100% sure about this one
     Wang~\textit{et al.}~\cite{wang2022disentangled} & 2022 & Face PAD & Rung 2 - T. R. P. & Live Features Disentanglement + t-SNE~\cite{JMLR:v9:vandermaaten08a}\\
     Sequeira~\textit{et al.}~\cite{sequeira2021exploratory} & 2021 & Face PAD & Rung 1 & Grad-CAM Visualisation\\
     Chen~\textit{et al.}~\cite{chen2021dual} & 2021 & 3D Mask Face PAD & Rung 1 & Grad-CAM Visualisation and t-SNE \\
     Neto~\textit{et al.}~\cite{neto2021myope} & 2021 & Face PAD & Rung 1 & Grad-CAM Visualisation \\
     Aghdaie~\textit{et al.}~\cite{aghdaie2021attention} & 2021 & Face PAD & Rung 2 - T. R. P. & Attention + Attention Maps Visualization \\ 
     Yu~\textit{et al.}~\cite{yu2020searching} & 2020 &  Face PAD & Rung 2 - M. P. + T. R. P. & Central Difference Convolution + Depth-Map Estimation \\
     Pinto~\textit{et al.}~\cite{pinto2020leveraging} & 2020 & Face PAD & Rung 1 & Grad-CAM Visualisation \\
     Wang~\textit{et al.}~\cite{wang2020deep} & 2020 &  Face PAD & Rung 2 - T. R. P. & Temporal Depth-Map Estimation + t-SNE \\
     Shao~\textit{et al.}~\cite{shao2020regularized} & 2020 & Face PAD & Rung 2 - T. R. P. & Meta-learning with Depth Regularization + Attention Map Visualization \\
    Liu~\textit{et al.}~\cite{liu2020disentangling} & 2020 & Face PAD & Rung 2 - T. R. P. & Spoof-traces disentanglement \\
    Yang~\textit{et al.}~\cite{yang2019face} & 2019 & Face PAD & Rung 2 - T. R. P.  & Patch-based Attention + Grad-CAM + t-SNE \\
    Jourabloo~\textit{et al.}~\cite{jourabloo2018face} & 2018 & Face PAD & Rung 2 - T. R. P. & Spoof Noise Modeling  + t-SNE \\
    Liu~\textit{et al.}~\cite{liu2018learning} & 2018 & Face PAD & Rung 2 - T. R. P. & rPPG Signal Extraction \\

    \hline

    \multicolumn{5}{|c|}{\makecell{\textbf{Face DeepFakes}}} \\
    \hline
    Mathews~\textit{et al.}~\cite{mathews2023explainable} & 2023 & Face DeepFake Detection & Rung 1 &  Grad-CAM Visualisations\\
    Silva~\textit{et al.}~\cite{silva2022deepfake} & 2022 & Face DeepFake Detection & Rung 1 &  Attention Maps\\
    Mazaheri and Roy-Chowdhury~\cite{mazaheri2022detection} & 2022 & Face Manipulation Detection & Rung 1 & CAM \\ % Is this deepFake detection?
    Korshunov~\textit{et al.}~\cite{korshunov2022custom} & 2022 & Face DeepFake Detection & Rung 1 &  t-SNE \\ %  used a multi-task approach for improved generalization and "explainability"
    Xu~\textit{et al.}~\cite{xu2022supervised} & 2022 & Face DeepFake Detection & Rung 1  & UMAP and Heatmap Visualisation \\

    \multicolumn{5}{|c|}{\makecell{\textbf{Face Emotion/Expression Recognition}}} \\
    \hline
    Nam~\textit{et al.}~\cite{nam4208671facialcuenet} & 2023 & Face Expression Recognition & Rung 2 - T. R. P. & Spatio-Temporal Attention Maps + Facial cues\\
    Rathod~\textit{et al.}~\cite{rathod2022kids} & 2022 & Children Emotion Recognition & Rung 1 & Grad-CAM + Grad-CAM++ + SoftGrad\\
    Cesarelli~\textit{et al.}~\cite{cesarelli2022emotion} & 2022 & Face Emotion Recognition & Rung 1 & Grad-CAM Visualisations\\
    Zhu~\textit{et al.}~\cite{zhu2022explainable} & 2022 & Face Emotion Recognition & Rung 1 & LRP\\
    Araf~\textit{et al.}~\cite{araf2022real} & 2022 & Face Emotion Recognition & Rung 1 & Grad-CAM Visualisations\\
    
    \multicolumn{5}{|c|}{\makecell{\textbf{Fingerprint PAD/Recognition}}} \\
    \hline
    Rai~\textit{et al.}~\cite{rai2023expressnet} & 2023 & Fingerprint PAD & Rung 1 &  Heatmap Visualisation\\
    Ramachandra and Li~\cite{ramachandra2023finger} & 2023 & Fingerphoto Verification & Rung 2 - I. I. & Grad-CAM + Occlusion Sensitivity Maps + LIME + Gradient Attribution\\
    Chowdhury~\textit{et al.}~\cite{chowdhury2020can} & 2020 & Fingerprint Recognition & Rung 1 & Grad-CAM Visualisation\\

    \multicolumn{5}{|c|}{\makecell{\textbf{Iris Recognition}}} \\
    \hline
    Boyd~\textit{et al.}~\cite{boyd2023human} & 2023 & Post-mortem Iris Recognition & Rung 2 - I. I. &  Visual Patch-matching\\
    Kuehlkamp~\textit{et al.}~\cite{kuehlkamp2022interpretable} & 2022 & Post-mortem Iris Recognition & Rung 1 &  Segmentation Masks + CAM Visualisations\\
    Hu~\textit{et al.}~\cite{hu2020end} & 2020 & Iris Recognition & Rung 1 & Grad-CAM Visualisation\\
    Trokielewicz~\textit{et al.}~\cite{trokielewicz2019perception} & 2018 & Cadaver Iris Recognition& Rung 1  & Grad-CAM Visualisation\\

    \multicolumn{5}{|c|}{\makecell{\textbf{Iris PAD}}} \\
    \hline
    Sharma and Ross~\cite{sharma2021viability} & 2021 & Iris PAD& Rung 1 & Grad-CAM Visualisation and t-SNE \\
    Fang~\textit{et al.}~\cite{fang2021iris} & 2021 & Iris PAD & Rung 2 - T. R. P. & Patch-based Supervision + Attention +  Score-Weighted CAM~\cite{zhang2021group}\\
    Chen and Ross~\cite{chen2021} & 2021 & Iris PAD & Rung 2 - T. R. P. & Channel and Position Attention +  Grad-CAM \\

    Sharma~\textit{et al.}~\cite{sharma2020d} & 2020 & Iris PAD & Rung 1 & Grad-CAM Visualisation and t-SNE \\
    Trokielewicz~\textit{et al.}~\cite{trokielewicz2018presentation} & 2018 & Cadaver Iris PAD & Rung 1 & Grad-CAM Visualisation and Guided Backpropagation  \\
    \hline

    \multicolumn{5}{|c|}{\makecell{\textbf{Others}}} \\
    \hline
    Diaz~\textit{et al.}~\cite{diaz4395568explainable} & 2023 & Signature Verification & Rung 2 - M. P. + T. P. & Universal Background Model(UBM) + Explainable Features \\
    Aquino~\textit{et al.}~\cite{aquino2022explaining} & 2022 & Accelerometer User Identification & Rung 1 & Grad-CAM\\
    Lim~\textit{et al.}~\cite{lim2022detecting} & 2022 & Voice DeepFake Detection & Rung 1 & Deep Taylor + LRP\\
    Algermissen and H{\"o}rnlein~\cite{algermissen2021person} & 2021 & Footstep Person Recognition & Rung 1 & Grad-CAM \\
    Alshazly~\textit{et al.}~\cite{alshazly2021towards} & 2021 & Ear Recognition & Rung 1 & Guided Grad-CAM \\
    
    Chandaliya and Nain~\cite{chandaliya2021child} & 2021 & Face Age Estimation & Rung 2 - T. R. P. & Attention \\ % Remove from the table? They don't even mention explainability
    Joshi~\textit{et al.}~\cite{Joshi2021} & 2021 & Fingerprint ROI Segmentation & Rung 2 - T. R. P. + I. P. &Monte Carlo Dropout + Uncertainty Map \\
    %\textcolor{orange}{Kao~\textit{et al.}~\cite{Kao2020AnOS}} & 2020 & Signature Verification & Rung 1 & - \\
    Pinto and Cardoso~\cite{pinto2020xecg} & 2020 & ECG Biometric Identification & Rung 1 & Gradient-SHAP + DeepLIFT + Saliency Maps\\
    Ahmed~\textit{et al.}~\cite{ahmed2020race} & 2020 & Ethnicity Estimation & Rung 1 & Grad-CAM \\
    Genovese~\textit{et al.}~\cite{Genovese2019} & 2019 & Face Aging & Rung 1 & Cross- GAN Filter Similarity Index \\
    %\hline
\bottomrule
    %\end{tabular}
\end{longtable}
\end{small}
\end{center}
\end{landscape}